%% file: main.tex
\documentclass{article}

\PassOptionsToPackage{numbers, compress}{natbib}
 \usepackage[preprint]{neurips_2026}

\usepackage[utf8]{inputenc} % allow utf-8 input
\usepackage[T1]{fontenc}    % use 8-bit T1 fonts
\usepackage{hyperref}       % hyperlinks
\usepackage{url}            % simple URL typesetting
\usepackage{booktabs}       % professional-quality tables
\usepackage{tabularx}       % full-width tables with auto-wrapping columns
\usepackage[section]{placeins} % keep floats within their section (auto \FloatBarrier at \section)
\usepackage{amsmath}        % \text in math mode, align, etc.
\usepackage{CJKutf8}        % Japanese/CJK characters in pdflatex
\usepackage{amsfonts}       % blackboard math symbols
\usepackage{nicefrac}       % compact symbols for 1/2, etc.
\usepackage{microtype}      % microtypography
\usepackage{xcolor}         % colors
\usepackage{colortbl}       % \rowcolor for table row banding
\usepackage{graphicx}       % \includegraphics for figures

\definecolor{HeaderBand}{HTML}{D6E4F0}      % light blue — header row
\definecolor{ProprietaryBand}{HTML}{E8EEF7} % light blue tint
\definecolor{OpenBand}{HTML}{E8F4EA}        % light green tint
\definecolor{MedicalBand}{HTML}{FCEAEA}     % light pink tint
\definecolor{MHLWBand}{HTML}{FFF1D6}        % warm peach — MHLW threshold row

\title{JMed48k: A Multi-Profession Japanese Medical Licensing Benchmark for Vision-Language Model Evaluation}

\author{%
  \textbf{Yue Xun}$^{1}$ \quad
  \textbf{Junyu Liu}$^{2}$ \quad
  \textbf{Qian Niu}$^{3,*}$ \quad
  \textbf{Xinyi Wang}$^{1}$ \quad
  \textbf{Zheng Yuan}$^{1}$ \quad
  \textbf{Zirui Li}$^{4}$ \\
  \textbf{Zequn Zhang}$^{5}$ \quad
  \textbf{Bowen Zhao}$^{6}$ \quad
  \textbf{Shujun Wang}$^{1}$ \quad
  \textbf{Irene Li}$^{3}$ \quad
  \textbf{Kan Hatakeyama-Sato}$^{3}$ \\
  \textbf{Yusuke Iwasawa}$^{3}$ \quad
  \textbf{Yutaka Matsuo}$^{3}$ \\
  $^{1}$The Hong Kong Polytechnic University \\
  $^{2}$Kyoto University \\
  $^{3}$The University of Tokyo \\
  $^{4}$Hohai University \\
  $^{5}$University of Science and Technology of China \\
  $^{6}$University of Toronto \\
  $^{*}$Corresponding author \\
  \texttt{qian.niu@weblab.t.u-tokyo.ac.jp}
}

\begin{document}

% Page-1 layout repair: NeurIPS' default \@notice uses a [b] float for
% the "Submitted to..." line, and that float reserves bottomfraction
% (40%) of page 1, blocking Section 1 from flowing onto page 1 even
% when the abstract is short.  Suppress the float and place the notice
% text manually at the page foot via a custom \thispagestyle.  The
% NeurIPS .sty file itself is untouched.
\makeatletter
\renewcommand{\@notice}{}
\def\ps@firstpage{%
  \let\@oddhead\@empty
  \let\@evenhead\@empty
  \def\@oddfoot{\hfil\footnotesize\@noticestring\hfil}%
  \def\@evenfoot{\hfil\footnotesize\@noticestring\hfil}%
}
\makeatother

\maketitle
\thispagestyle{firstpage}

\begin{abstract}
We introduce JMed48k, a multi-profession Japanese healthcare licensing benchmark for evaluating vision-language models. Built from official PDF materials released by the Japanese Ministry of Health, Labour and Welfare, JMed48k contains 48{,}862 exam questions and 20{,}142 images from 11 national licensing examinations between 2005 and 2025, with visual content annotated under an 8-type taxonomy. From this corpus, we derive JMed48k-Eval, a recent five-year evaluation subset with 12{,}484 scored questions, including 9{,}905 text-only questions and 2{,}579 questions with images. We evaluate 21 proprietary, open-source, and medical-specific models, reporting text-only and with-image performance separately. Because these subsets contain different questions, we further introduce a paired image-removal audit that evaluates questions with images before and after removing visual content to explore four answer-transition states. The audit shows that proprietary and open-source models gain substantially from images, whereas medical-specific systems show limited observable use of visual evidence, with many correct answers persisting after image removal. Even among proprietary models, the net image-removal effect varies sevenfold across professions, from +5.7 points on Physician questions to +39.8 points on Public Health Nurse questions. We release JMed48k to support reproducible, profession-stratified evaluation of vision-language models in medical licensing settings.
\end{abstract}

%======================================================================
% \section{Introduction}\label{sec:intro}
%======================================================================

\section{Introduction}\label{sec:intro}

\begin{figure}[!tbp]
  \centering
  \includegraphics[width=\linewidth]{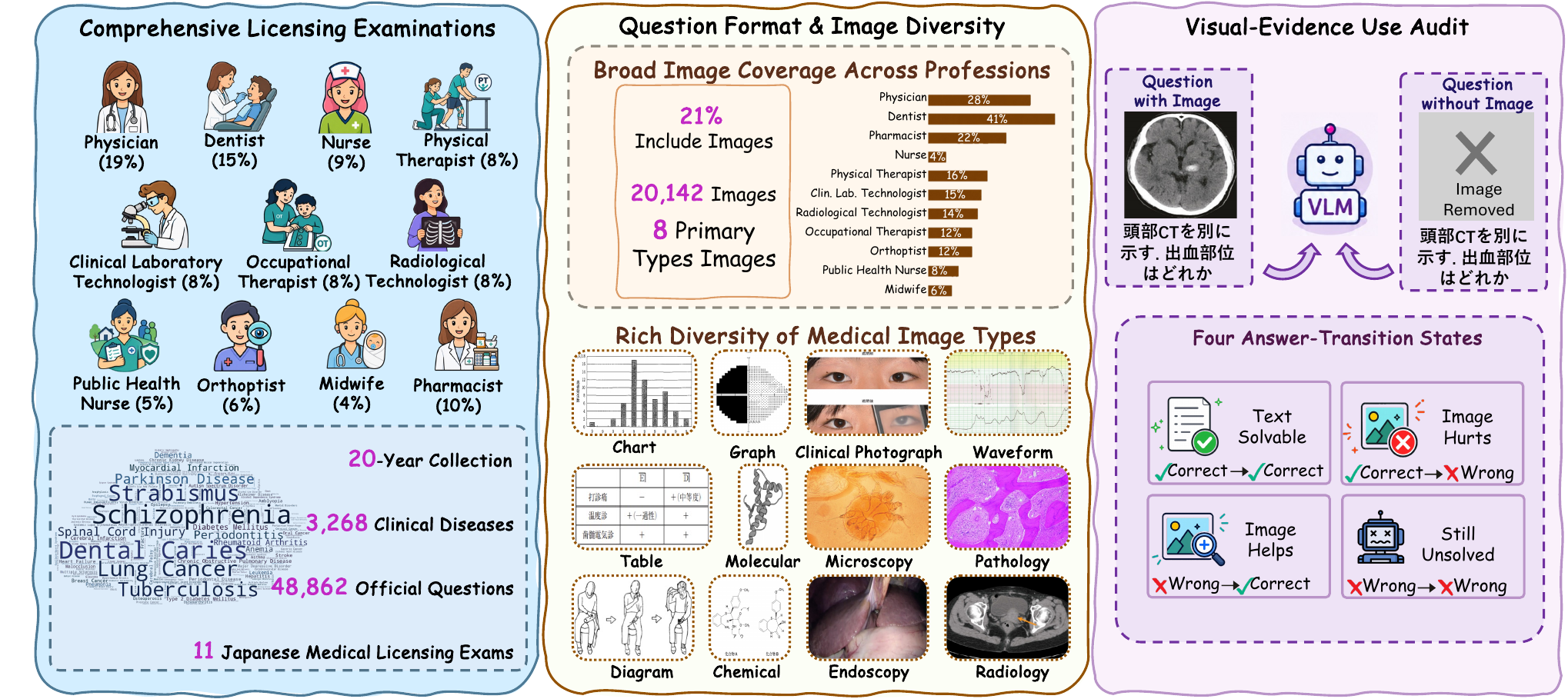}
  \caption{Overview of JMed48k benchmark. Left: JMed48k aggregates 48,862 official questions from 11 Japanese national healthcare licensing examinations spanning 2005–2025. Middle: Questions with images span all professions and account for 21\% of the corpus, with visual content annotated under an 8-type taxonomy. Right: The paired image-removal audit re-evaluates each question with images in JMed48k-Eval by removing all images, revealing four distinct outcomes: text-solvable success, image-enabled correction, image-induced error, and still-unsolved items. }
  \label{fig:intro-overview}
\end{figure}

\input{tables/T1a_text_only}

% Vision language models are increasingly evaluated on medical question answering and clinical reasoning tasks. PubMedQA \citep{jin2019pubmedqa} established biomedical research question answering as a reusable evaluation task. MultiMedQA and Med PaLM \citep{singhal2023clinical} connected professional medical exams, biomedical research questions, and consumer health questions in a unified medical question answering setting. Med-PaLM2 \citep{singhal2025expert}, the ChatGPT USMLE evaluation \citep{kung2023chatgpt_usmle}, and the GPT 4 medical challenge evaluation \citep{nori2023gpt4_medical} further showed that modern language models can reach strong performance on medical exam style questions. Professional licensing examinations are therefore an attractive source for evaluating medical AI. They are written and validated by domain experts, have official answer keys, and are designed to test the knowledge required for entry into regulated clinical practice.

Medical AI evaluation has increasingly used question answering and professional examinations to measure clinical knowledge and reasoning. PubMedQA \citep{jin2019pubmedqa} established biomedical research question answering as a reusable evaluation task. MultiMedQA and MedPaLM \citep{singhal2023clinical} connected professional medical exams, biomedical research questions, and consumer health questions within a unified medical question answering framework. MedPaLM2 \citep{singhal2025expert}, the ChatGPT USMLE evaluation \citep{kung2023chatgpt_usmle}, and the GPT 4 medical challenge evaluation \citep{nori2023gpt4_medical} further showed that modern language models can perform strongly on medical exam style questions. For vision language models, licensing examinations offer a natural next step because many medical exam items require models to integrate textual stems with visual evidence while retaining the authority of expert written questions and official answer keys.

For vision language models, licensing examinations offer an additional advantage. Many medical exam questions require candidates to combine a textual stem with visual evidence such as radiographs, clinical photographs, pathology images, waveforms, charts, and chemical structures. These items test whether a model can connect visual observations with clinical knowledge rather than relying only on memorized text. A medical licensing benchmark with broad professional coverage can therefore evaluate both medical knowledge and visual evidence use in a setting that is grounded in real credentialing practice.

Existing medical exam benchmarks cover only part of this setting. MedQA \citep{jin2021medqa} focuses on English medical licensing questions. MedMCQA \citep{pal2022medmcqa} draws from Indian medical entrance examinations. MLECQA \citep{li2021mlecqa} and CMExam \citep{liu2023cmexam} cover Chinese medical examination questions. KorMedMCQA \citep{kormedmcqa} targets Korean healthcare professional licensing examinations. IgakuQA \citep{igakuqa} evaluates Japanese physician licensing questions. YakugakuQA \citep{yakugakuqa} evaluates Japanese pharmacist licensing questions. These resources have enabled medical exam evaluation across languages, but they are primarily text based or centered on one profession. WorldMedQA-V \citep{worldmedqav} and KorMedMCQA-V \citep{kormedmcqav} extend medical licensing evaluation to image bearing questions, but they still focus on physician oriented settings. KokushiMD-10 \citep{kokushimd10} broadens Japanese evaluation to multiple healthcare professions, but it covers fewer professions and a narrower temporal window than the full national licensing archive.

This limited coverage matters because healthcare is delivered by a broad workforce. Japanese national healthcare licensing examinations cover physicians, dentists, pharmacists, nurses, public health nurses, midwives, radiological technologists, clinical laboratory technologists, physical therapists, occupational therapists, and orthoptists. These professions differ in clinical focus, vocabulary, exam structure, and the types of visual evidence. A model that performs well on physician questions may not show the same capability on pharmacy charts, nursing public health tables, dental photographs, rehabilitation diagrams, or laboratory microscopy. Whether vision language model performance generalizes across this full professional spectrum remains largely unknown.

A second gap concerns how multimodal ability is measured. VQA-RAD \citep{vqarad}, SLAKE \citep{slake}, PathVQA \citep{pathvqa}, PMCVQA \citep{zhang2023pmcvqa}, OmniMedVQA \citep{hu2024omnimedvqa}, GMAI-MMBench \citep{chen2024gmai}, MultiMedEval \citep{royer2024multimedeval}, and MMMU \citep{yue2024mmmu} evaluate medical or broad expert level multimodal question answering. These benchmarks usually report accuracy on image-bearing subsets, sometimes with breakdowns by modality, task, or department. VQA-v2 \citep{goyal2017vqav2} and VQA-CP \citep{agrawal2018vqacp} showed that visual question answering systems can exploit language priors. POPE \citep{li2023pope}, HallusionBench \citep{guan2024hallusionbench}, and MMVP \citep{tong2024mmvp} further test visual illusion, and visual perception failures in general vision language models. DrVDBench \citep{drvdbench} introduces medical erasure tasks that test whether models rely on visible image regions. These efforts are important, but aggregate accuracy and net ablation scores still hide how individual answers change when images are present or removed. The distinction is important. A positive image effect can mean that images consistently turn wrong answers into correct answers. It can also mean that images help on some questions while harming others by a similar amount. A correct answer with images can reflect genuine visual reasoning, but it can also reflect a question that is solvable from text alone. To identify whether images are used, ignored, or misleading, evaluation must track paired answer transitions at the item level.

We introduce JMed48k to address these gaps. JMed48k is a multi profession Japanese medical licensing benchmark constructed from official examination materials released by the Japanese Ministry of Health, Labour and Welfare (MHLW). The full corpus contains 48,862 questions from 11 national healthcare licensing examinations from 2005 to 2025. Among these questions, 9,646 reference 20,142 images annotated under an 8 type visual content taxonomy. All experiments use JMed48k-Eval, a recent five-year subset of the corpus with 12{,}484 scored questions, including 9{,}905 text-only questions and 2{,}579 questions with images.

Together with the dataset, we introduce a paired image removal audit. For every question with images, a model is evaluated twice on the same textual input. One run includes the original images, and the other removes all visual content. The paired outcomes assign each question to one of four states. The answer may be correct in both settings, correct only with images, correct only without images, or incorrect in both settings. This design separates text solvable success, image enabled correction, image induced error, and persistent failure.

We evaluate 21 models spanning proprietary systems, open source general purpose systems, and medical specific systems. The paired audit is applied to all 20 models that accept images. The results show that proprietary and open source systems gain substantially from images, while medical specific systems show limited observable use of visual evidence. Many correct answers from medical specific systems remain correct after the images are removed. The importance of images also varies sharply by profession. For proprietary models, removing images reduces accuracy by only 5.7 points on Physician questions but by 39.8 points on Public Health Nurse questions. This sevenfold spread shows that conclusions drawn from one profession or one pooled image subset transfer poorly to the broader healthcare workforce. Our contributions are as follows:
\begin{itemize}
\item \textbf{JMed48k corpus.} A multi-profession medical VLM benchmark containing 48{,}862 questions from 11 Japanese national healthcare licensing examinations across 2005--2025, of which 9{,}646 include images annotated under an 8-type content taxonomy.
\item \textbf{Systematic evaluation of 21 models across three families.} Including 6 proprietary, 7 open-source general-purpose, and 8 medical-specific models, revealing cross-profession performance variation and exposing a gap between aggregate accuracy and audited image use that standard reporting obscures.
\item \textbf{Paired image-removal audit.} A paired image-removal audit that evaluates each question with and without visual evidence, decomposing performance into four answer-transition states and enabling fine-grained analysis of visual reliance across professions, image types, and model families.
\end{itemize}

\section{Related Work}\label{sec:related}

\subsection{Medical Licensing Examination Benchmarks}

Medical licensing examinations provide expert written questions and official answer keys, making them natural benchmarks for clinical reasoning. Existing resources span several languages, including MedQA \citep{jin2021medqa} for English physician licensing questions, MedMCQA \citep{pal2022medmcqa} for Indian medical entrance questions, MLECQA \citep{li2021mlecqa} and CMExam \citep{liu2023cmexam} for Chinese medical examinations, and KorMedMCQA \citep{kormedmcqa} for Korean healthcare professional licensing examinations. Model focused studies such as the ChatGPT USMLE evaluation \citep{kung2023chatgpt_usmle} and the GPT 4 medical challenge evaluation \citep{nori2023gpt4_medical} further show that licensing-based questions have become a standard testbed for medical reasoning. However, this line of work remains largely text based and physician oriented or single profession.

Japanese and multimodal licensing resources are more recent. IgakuQA \citep{igakuqa} evaluates Japanese physician licensing questions. JPharmaBench \citep{yakugakuqa} introduces YakugakuQA for Japanese pharmacist licensing questions. JMedBench \citep{jiang2025jmedbench} covers broader Japanese biomedical tasks, but is not a licensing examination benchmark. WorldMedQA-V \citep{worldmedqav} and KorMedMCQA-V \citep{kormedmcqav} add image bearing licensing questions, but remain focused on physician oriented settings. Closest to JMed48k is KokushiMD-10 \citep{kokushimd10}, which assembles Japanese national licensing examinations with text and image items. JMed48k extends this line by using official MHLW materials, covering all 11 Japanese national healthcare licensing professions from 2005 to 2025, separating text only and with image evaluation, and applying paired image removal to every image bearing question.

\input{tables/T1b_with_images}

\subsection{Multimodal Medical Benchmarks}

Medical multimodal evaluation has moved from focused VQA datasets to broader clinical benchmark suites. VQA-RAD \citep{vqarad} and VQA-Med \citep{benabacha2019vqamed} established early medical visual question answering settings. SLAKE \citep{slake} added semantic labels and medical knowledge structure. PathVQA \citep{pathvqa} and PMC-VQA \citep{zhang2023pmcvqa} expanded evaluation to pathology and PubMed Central figures. More recent benchmarks increase clinical breadth. PathMMU \citep{pathmmu} targets expert level pathology reasoning. OmniMedVQA \citep{hu2024omnimedvqa} spans twelve imaging modalities. GMAI-MMBench \citep{chen2024gmai} organizes evaluation by clinical tasks, departments, modalities, and perceptual granularity. MultiMedEval \citep{royer2024multimedeval} provides a unified medical VLM evaluation toolkit. TCM-Ladder \citep{tcmladder} extends multimodal evaluation to traditional Chinese medicine.

Medical VLM development has progressed in parallel, with LLaVA-Med \citep{li2023llavamed}, Med Flamingo \citep{moor2023medflamingo}, RadFM \citep{wu2023radfm}, and HuatuoGPT Vision \citep{chen2024huatuogptvision} adapting general multimodal modeling to biomedical figures, radiology data, medical image text pairs, and visual instruction data. General multimodal benchmarks such as MME \citep{fu2023mme}, MMBench \citep{liu2023mmbench}, ScienceQA \citep{lu2022scienceqa}, SEEDBench \citep{li2024seedbench}, MMVet \citep{yu2024mmvet}, and MMMU \citep{yue2024mmmu} provide complementary tests of perception, reasoning, and integrated multimodal capability. However, these benchmarks are not organized around professional medical licensing and do not test generalization across a national healthcare workforce.

A related line of work studies whether models rely on images or textual shortcuts. VQA-v2 \citep{goyal2017vqav2} and VQA-CP \citep{agrawal2018vqacp} exposed language priors in visual question answering. POPE \citep{li2023pope}, HallusionBench \citep{guan2024hallusionbench}, and MMVP \citep{tong2024mmvp} study hallucination and visual perception failures. DrVDBench \citep{drvdbench} brings visual reliance testing into medical imaging through organ erasure and lesion erasure tasks. JMed48k is complementary to these efforts because it evaluates the same licensing question with and without all visual evidence, separating text solvable success, image enabled correction, image induced error, and persistent failure across 11 healthcare professions.

\section{The JMed48k Benchmark}\label{sec:dataset}

\subsection{Source examinations and corpus}\label{sec:dataset-source}

JMed48k is constructed from official examination materials released by the Japanese MHLW, which administers 11 national healthcare professional licensing examinations, covering Physician, Dentist, Pharmacist, Nurse, Public Health Nurse, Midwife, Physical Therapist, Occupational Therapist, Orthoptist, Radiological Technologist, and Clinical Laboratory Technologist. These examinations serve as the legal entry point to their corresponding professions. Figure~\ref{fig:intro-overview} summarizes the corpus scale, profession coverage, visual taxonomy, and a representative question with images.

The \textbf{full JMed48k} corpus aggregates questions released by MHLW across these 11 examinations from exam years 2005--2025. It contains $48{,}862$ official questions, of which $9{,}646$ reference one or more medical images. From this corpus, we derive \textbf{JMed48k-Eval}, the evaluation subset used throughout this paper, restricted to the most recent five years available for each profession. Eight examinations cover 2021--2025, while Public Health Nurse, Midwife, and Nurse cover 2020--2024 because usable 2025 examination PDFs were not available at the time of construction. After excluding $93$ items for which MHLW withdrew the question or did not publish a usable official gold answer (Appendix~\ref{app:curation}), JMed48k-Eval contains $12{,}484$ scored questions across 55 profession-year cells, comprising $9{,}905$ text-only questions and $2{,}579$ questions with images. All accuracies reported in this paper are computed on this scored subset. Appendix~\ref{app:profession_breakdown} and Table~\ref{tab:profession_breakdown} report the full per-profession breakdown of JMed48k-Eval.

\subsection{Construction and quality control}\label{sec:dataset-construction}

For each profession-year, we convert the question, figure, and answer PDFs into one structured JSON record per question. MinerU~\citep{wang2024mineru} parses the question PDF to extract stems, answer choices, and references to visual materials. The figure PDF provides the referenced images, which are cropped and matched by section and question-number captions. The answer PDF provides official gold answers, which are joined by question identifier.

Quality control combines professional Japanese-language manual verification with automated consistency checks. The manual review compares extracted records against the source PDFs, covering OCR artifacts, option order, chemical and isotope notation, image extraction, image paths, and image-question pairing. Scripts check missing fields, duplicated identifiers, malformed option keys, missing image files, and metadata inconsistencies. JMed48k preserves official answer structures, including single-choice answers, multiple correct-option answers, alternative acceptable answer sets, and numeric-slot answers. PDF processing details, correction logs, schema examples, and scoring rules are provided in Appendices~\ref{app:curation}--\ref{app:scoring}.

\subsection{Visual content and dataset composition}\label{sec:dataset-taxonomy}

Questions with images are annotated using an 8-type visual-content taxonomy: Clinical Photograph, Radiology, Pathology/Microscopy, Endoscopy, Chart/Graph/Table, Diagram, Chemical/Molecular, and Waveform/Signal. A small Other/Unclear bucket is retained for image references whose visual content MHLW does not publicly redistribute (typically for privacy or rights reasons such as patient portraits); these records preserve the existence of an image reference but carry no inspectable image and are excluded from type-stratified analyses. Labels are assigned at the image level, since a question may reference multiple images, and aggregated to the question level for analyses. The full corpus contains 9{,}646 questions with images and 20{,}142 image-level annotations. Appendix~\ref{app:image_taxonomy} describes the annotation procedure, and Appendix Table~\ref{tab:dataset_image_types} reports the image-type distribution.

JMed48k inherits substantial heterogeneity across professions. In the scored JMed48k-Eval subset, the share of questions with images ranges from 4.3\% for Nurse to 43.8\% for Dentist, with an overall share of 20.8\%. 
Professions also differ in the distribution of these image types, which motivates profession-stratified and image-type-stratified evaluation rather than a single pooled score.

JMed48k records the metadata needed for paired image removal. Each question with images stores its linked image files, the image labels attached to the question, and a flag for cases where the answer choices themselves are images. These fields allow the same question to be evaluated with and without visual content, while supporting analyses by profession, image type, and visual-option format. Appendix~\ref{app:profession_breakdown} reports the full profession-level composition, including dominant image type, image-as-options counts, and answer-format counts.

%======================================================================
\section{Experiments}\label{sec:results}

Our experiments are organized as two baselines followed by a paired visual evidence audit. 
Sections~\ref{sec:text-only} and~\ref{sec:image-bearing} report accuracy on text only and 
image bearing questions separately, since the two subsets contain different items and test 
different capabilities. Sections~\ref{sec:image-removal} and~\ref{sec:profession-image-use} 
then apply paired image removal to questions with images, analyzing answer transitions 
overall, by model family, and by profession. This structure distinguishes ordinary accuracy 
from observable use of visual evidence.s Inference settings, prompts, and additional setup details are provided in Appendix~\ref{app:setup}.

We evaluate 21 models in three cohorts. Proprietary models include Gemini 2.5 Flash and Pro \citep{comanici2025gemini25}, GPT-5 mini and GPT-5 \citep{openai2025gpt5}, Claude Sonnet 4 \citep{anthropic2025claude4}, and Grok 4.20 \citep{xai2025grok4}. Open-source general-purpose models include Qwen3.5-VL 9B, 27B, and 397B-A17B with the closest verified Qwen3-VL family provenance \citep{bai2025qwen3vl}, Gemma 4 IT \citep{google2026gemma4}, Llama 4 Maverick \citep{meta2025llama4}, GLM-4.6V \citep{zai2025glm46v} with GLM-V foundation \citep{hong2025glm45v}, and DeepSeek-R1 \citep{guo2025deepseekr1}. Medical-specific models include Lingshu \citep{xu2025lingshu}, MedGemma \citep{sellergren2025medgemma}, and Hulu-Med \citep{jiang2025hulumed}. DeepSeek-R1 is a text-only modal excluded from image-removal analyses.

\begin{figure}[!t]
  \centering
  \includegraphics[width=0.96\linewidth]{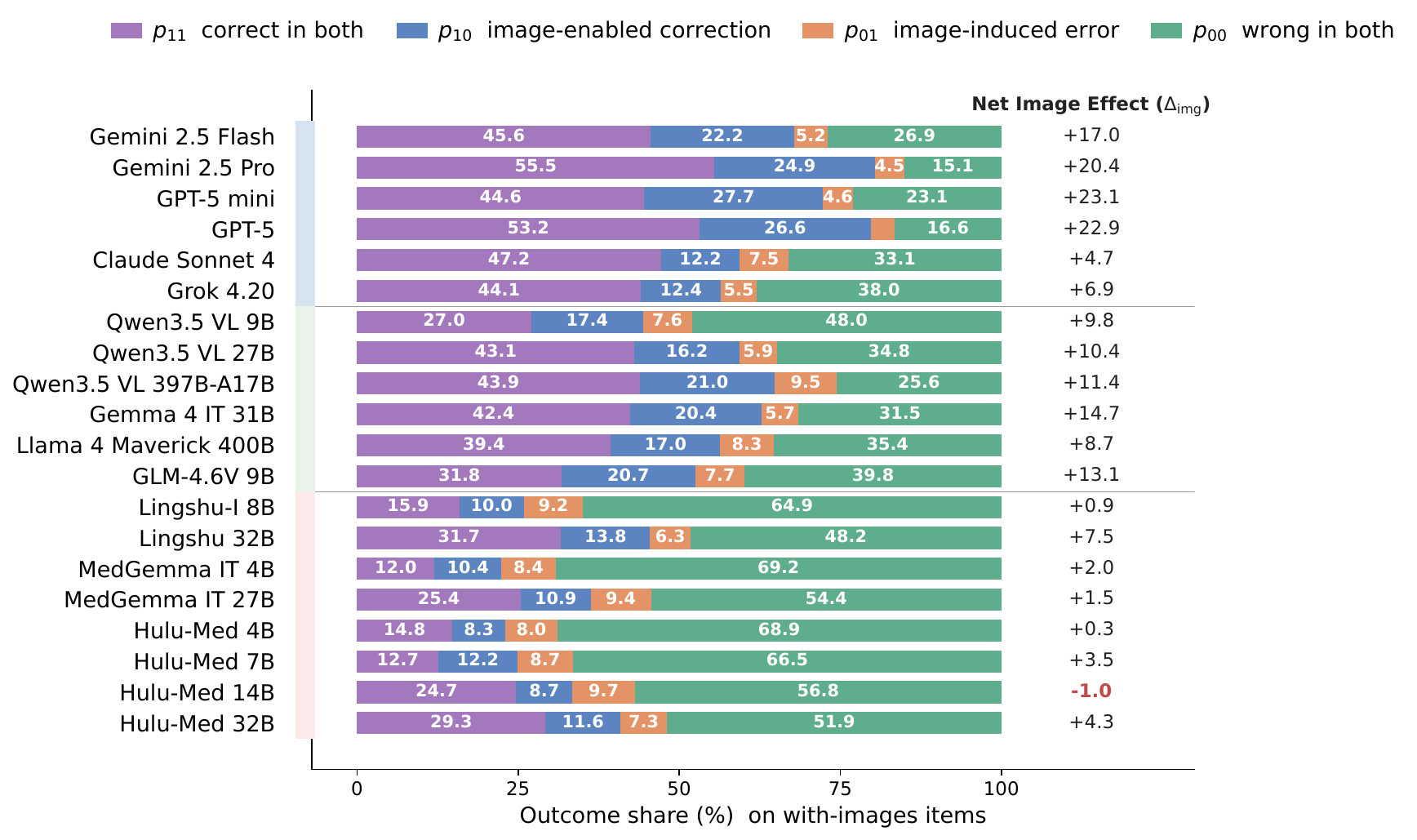}
  \caption{\textbf{Paired image-removal audit.} Each row decomposes a
  model's predictions on the 2{,}579 questions with images into four
  paired outcomes: correct with and without images ($p_{11}$), correct
  only with images ($p_{10}$), correct only after image removal
  ($p_{01}$), and incorrect in both settings ($p_{00}$). The right
  column reports the net image effect, with negative values shown in
  red. Per-model counts are provided in
  Appendix~\ref{app:transition-details}.}
  \label{fig:image-removal-transitions}
\end{figure}

\subsection{Text-only questions across professions}\label{sec:text-only}
Table~\ref{tab:t1a} reports text-only accuracy on the 9{,}905-item
text-only subset across all 11 professions.
Models separate into three tiers with large gaps between them.
The proprietary tier clears 95\% on the macro average, led by Gemini 2.5 Pro at 95.5\% and GPT-5 at 95.1\%,
followed by Claude~Sonnet~4 at 89.3\% and the GPT-5 mini and
Gemini~2.5 Flash variants in the high 80s.
The open-source tier is led by Qwen3.5~VL~397B-A17B at 86.5\% and
Gemma~4~IT~31B at 84.4\%, with the smallest Qwen3.5~VL~9B at 68.9\%
forming the lower bound.

The medical-specific tier falls substantially below the open-source floor. The strongest medical-specific model Lingshu~32B reaches only 65.0\%, about 17\% below the weakest proprietary model Grok 4.20, and the remaining seven score between 27.7\% and 60.6\%.

Per-profession scores vary substantially even within the same
model, ranging from 13\% to 26\% between the easiest and hardest
profession for high-performing models.
Dentistry, Orthoptist, and Public Health Nurse items are
consistently the hardest across all three tiers, while Physician
and Nurse items are consistently the easiest.
Lingshu~32B scores at or below the open-source tier on all 11
professions, indicating that medical-specific instruction tuning
in its current form does not transfer broadly across Japanese
licensing professions.

\subsection{Questions with images across professions}\label{sec:image-bearing}

Table~\ref{tab:t1b} reports model accuracy on the with-images subset. Performance on these items falls substantially below the text-only baseline. However, the magnitude of this accuracy drop is highly uneven. Among proprietary models with similar text-only ceilings, the gap ranges from 15.8\% for Gemini~2.5~Pro to over 28\% for Claude~Sonnet~4. The profession axis introduces even greater variance. The mean performance drop spans from 8.3\% on Physician items to 30.8\% on Occupational Therapist items. The profession-level spread exceeds the cross-model spread among proprietary systems, which motivates evaluating each licensing exam separately rather than reporting a single pooled score. The full per-(model, profession) accuracy-gap matrix is provided in Appendix~\ref{app:subset_gap_matrix}.

Crucially, the with-images subset contains different questions from the text-only baseline. A lower aggregate score on this subset merely reflects differing question difficulty and topic distribution rather than proving that visual input impairs reasoning. To isolate whether the images themselves help or harm performance, we must evaluate identical items with and without their visual evidence. We conduct this controlled within-item audit in the following sections.

\begin{figure}[t]
  \centering
  \includegraphics[width=\linewidth]{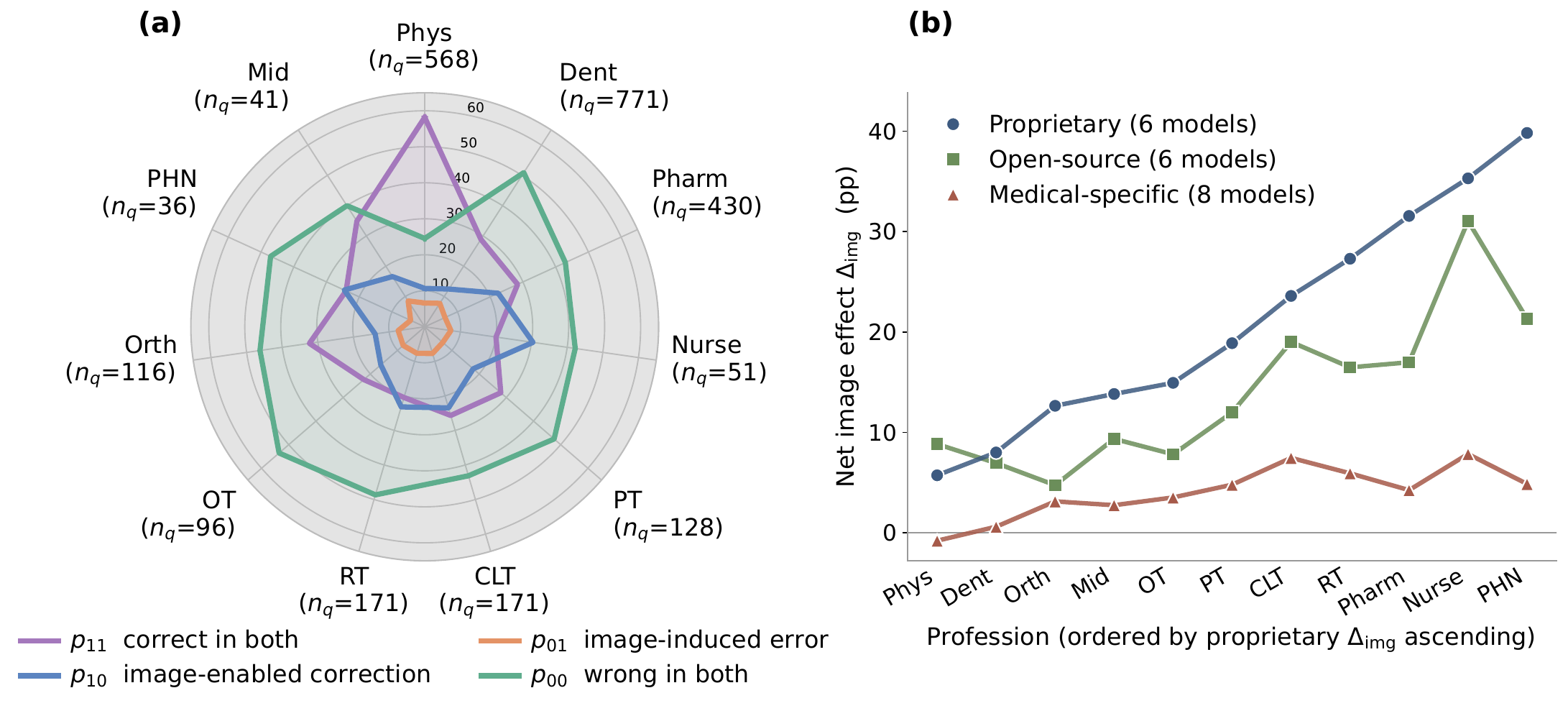}
    \caption{\textbf{Profession-level visual reliance.} 
    (a) shows the four paired answer-transition states by profession, 
    pooled over the 20 multimodal-capable models. State $p_{11}$ counts 
    items answered correctly in both settings, $p_{10}$ those answered 
    correctly only with the image, $p_{01}$ those answered correctly 
    only after image removal, and $p_{00}$ those answered incorrectly 
    in both. Per-profession item counts $n_q$ are also reported. 
    (b) shows the cohort-stratified profession-level net image effect 
    $\Delta_{\mathrm{img}} = a_{\mathrm{with}} - a_{\mathrm{removed}}$, 
    with professions ordered by the proprietary cohort's 
    $\Delta_{\mathrm{img}}$ ascending. Profession abbreviations are 
    Phys for Physician, Dent for Dentist, Pharm for Pharmacis, PT for Physical Therapist, CLT for Clinical 
    Laboratory Technologist, RT for Radiological Technologist, 
    OT for Occupational Therapist, Orth for Orthoptist, 
    PHN for Public Health Nurse, and Mid for Midwife.}
  \label{fig:profession-combined}
\end{figure}

\subsection{Auditing visual evidence use}\label{sec:image-removal}
The text-only and with-images subsets contain different questions. The accuracy gap between them therefore mixes image effects with differences in item content and cannot identify the image as the source of difficulty. To isolate the contribution of visual evidence, we re-evaluate every question with images after removing all attached images while keeping the question texts unchanged.

Comparing the two correctness records partitions every question into one of four answer-transition states, illustrated in Figure~\ref{fig:image-removal-transitions}. We use \(p_{11}\) for questions answered correctly in both settings, \(p_{10}\) for questions answered correctly only with the image, \(p_{01}\) for questions answered correctly only after image removal, and \(p_{00}\) for questions answered incorrectly in both settings. The original-setting accuracy is \(a_{\mathrm{with}}=p_{11}+p_{10}\), the image-removed accuracy is \(a_{\mathrm{removed}}=p_{11}+p_{01}\), and the net image effect is \(\Delta_{\mathrm{img}}=a_{\mathrm{with}}-a_{\mathrm{removed}}=p_{10}-p_{01}\). While \(\Delta_{\mathrm{img}}\) provides a high-level summary, it can hide substantial answer movement. For example, \(\Delta_{\mathrm{img}}\) of Hulu-Med 4B is only \(+0.3\) points, but its \(p_{10}\) and \(p_{01}\) are \(8.3\%\) and \(8.0\%\), respectively. Thus, image removal changes the correctness outcome on \(16.3\%\) of paired questions, even though gains and losses nearly cancel in the aggregate.

Three patterns emerge across the 20 multimodal models. The
proprietary models record large $p_{10}$ against small
$p_{01}$. The GPT-5 and Gemini~2.5 families achieve $p_{10}$
between 22\% and 28\% against $p_{01}$ under 6\%. Image input
shifts substantially more answers from wrong to right than the
reverse and yields large positive $\Delta_{\mathrm{img}}$, reaching
$+23.1$ points for GPT-5~mini.\footnote{A small fraction of GPT-5
family responses under image removal are calibrated refusals rather
than parse failures, and a forced-guess probe confirms that these
refusals reflect genuine epistemic uncertainty. Full details and
alternative $\Delta_{\mathrm{img}}$ variants appear in
Appendix~\ref{app:refusal}.} Open-source models such as
Qwen3.5~VL~397B-A17B, Gemma~4~IT~31B, and GLM-4.6V~9B show similar
behavior at a smaller magnitude, with $p_{10}$ near 20\% and
$p_{01}$ between 7\% and 10\%. Medical-specific models exhibit a different pattern. Their
$p_{10}$ and $p_{01}$ values are similar in magnitude, with
$p_{10}$ between 8\% and 14\% and $p_{01}$ between 6\% and 10\%,
so $\Delta_{\mathrm{img}}$ clusters near zero across the cohort.
Their with-images accuracy rests largely on $p_{11}$ rather than
on image-driven correction. Hulu-Med~14B is the single model in
our suite where image input is net harmful, with $p_{01}$ at 9.7\%
exceeding $p_{10}$ at 8.7\%.

A direct test on items where the answer choices are visually
embedded rather than enumerated in text, reported in
Appendix~\ref{app:image_options}, finds that the medical-specific
cohort scores at random on these items, suggesting that their
$p_{10}$ and $p_{01}$ in the paired audit reflect text-prior
behavior rather than image-driven correction. Four representative
answer-transition examples drawn from Physician 2021, illustrating
each of the four states, appear in Appendix~\ref{app:cases}.

% \subsection{Profession-level structure of image use}\label{sec:profession-image-use}

\subsection{Image reliance varies by profession}\label{sec:profession-image-use}

% \begin{figure}[t]
%   \centering
%   \includegraphics[width=\linewidth]{figures/fig_profession_combined.pdf}
%   \caption{\textbf{Profession-level visual reliance.}
%   \textbf{(a)} Pooled over the 20 multimodal models, each profession's
%   with-images items are decomposed into the four paired
%   answer-transition states.  Profession-level item counts $n_q$ are
%   shown in parentheses.
%   \textbf{(b)} Cohort-stratified profession-level net image effect
%   $\Delta_{\mathrm{img}} = a_{\mathrm{with}} - a_{\mathrm{removed}}$,
%   with professions ordered by the proprietary cohort's
%   $\Delta_{\mathrm{img}}$ ascending.
%   Profession abbreviations: Phys = Physician, Dent = Dentist,
%   Pharm = Pharmacist, Nurse = Nurse, PT = Physical Therapist,
%   CLT = Clinical Laboratory Technologist, RT = Radiological
%   Technologist, OT = Occupational Therapist, Orth = Orthoptist,
%   PHN = Public Health Nurse, Mid = Midwife.}
%   \label{fig:profession-combined}
% \end{figure}

\paragraph{Profession-level transition composition.}

Pooling over the 20 multimodal-capable models, Figure~\ref{fig:profession-combined}(a) decomposes each profession's with-images items into the four paired transition states. Two regimes are visible. Physician items are dominated by text-solvable success, with $p_{11}$ at $58.2\%$ and image-enabled correction $p_{10}$ at only $10.7\%$. Nurse items invert this balance, with $p_{10}$ at $30.4\%$ and $p_{11}$ at $20.0\%$, while Public Health Nurse sits at the crossover where the two states are nearly equal at $24.0\%$ and $24.6\%$. Image-induced errors $p_{01}$ never exceed $8.5\%$ on any profession, so cross-profession variation in $\Delta_{\mathrm{img}}$ is driven by how much $p_{10}$ shifts the answer from wrong to right rather than by images flipping correct text-only answers wrong.

\paragraph{Image contribution varies sevenfold across professions.}

The polarization is preserved when we split by model family. Figure~\ref{fig:profession-combined}(b) shows the proprietary cohort's $\Delta_{\mathrm{img}}$ rising from $+5.7$ on Physician and $+8.0$ on Dentist to $+27.3$ on Radiological Technologist, $+31.6$ on Pharmacist, $+35.3$ on Nurse, and $+39.8$ on Public Health Nurse, a sevenfold spread across professions. Open-source models trace the same gradient at lower magnitude, peaking at $+31.0$ on Nurse and $+21.3$ on Public Health Nurse. Medical-specific models flatten the gradient entirely, with $\Delta_{\mathrm{img}}$ below $+8$ everywhere and turning slightly negative on Physician at $-0.8$. Per-cohort values are reported in Appendix Table~\ref{tab:profession_delta_full}.

\paragraph{Text recoverability follows image-type composition.}

The profession-level gradient tracks the image-type composition of each licensing exam. Physician items are $45.4\%$ Radiology/Scan and Dentist $64.5\%$ Clinical Photograph (Appendix Table~\ref{tab:profession_breakdown}). On these two types, proprietary $a_{\mathrm{removed}}$ already stands at $68.1\%$ and $55.1\%$, and the type-level $\Delta_{\mathrm{img}}$ is only $+9.2$ and $+7.6$ (Appendix Table~\ref{tab:image_type_delta_full}), so the stem alone typically carries enough clinical context for the answer. Pharmacist and Public Health Nurse load instead on Chart/Graph/Table and Chemical/Molecular content, accounting for $81.2\%$ and $97.2\%$ of their with-images items. On these types, the stem describes what the image asks about but not what it depicts, and the type-level $\Delta_{\mathrm{img}}$ rises to $+24.4$ and $+38.1$ in the proprietary cohort. The same with-images label therefore reflects very different degrees of multimodal demand across licensing professions.

%======================================================================
% \section{Conclusion}\label{sec:conclusion}
%======================================================================
% We introduce JMed48k, a multi-profession Japanese healthcare licensing benchmark collected from 11 national exams over two decades, with 48,862 questions, 9,646 of which include images. Separately reporting accuracy on text-only and image subsets reveals that image accuracy falls substantially below the text-only baseline across all 21 evaluated proprietary, open-source, and medical-specific models, and the drop varies sharply by profession. Because the two subsets contain different questions, we paired the dataset with an image-removal audit that re-evaluates every image question without visual content and classifies answer shifts into four transition states. The audit shows that even medical-specific systems fail to use visual evidence reliably, and the net image effect on proprietary models ranges from $+5.7$ points on Physician to $+39.8$ points on Public Health Nurse, a sevenfold spread that aggregate accuracy and net image effect hide. We publicly release the corpus, an image taxonomy, and the paired-audit protocol to support reproducible evaluation of visual evidence use in medical vision-language models.

\section{Conclusion}\label{sec:conclusion}

We introduce JMed48k, a multi profession Japanese healthcare licensing benchmark collected from 11 national examinations over two decades, with 48{,}862 questions, including 9{,}646 questions that reference images. All experiments use JMed48k-Eval, a recent five year subset with 12{,}484 questions. Separately reporting text only and image bearing accuracy shows that image bearing questions remain substantially harder across proprietary, open source, and medical specific models. Our paired image removal audit classifies each image bearing item into four transition states, showing limited observable visual evidence use in medical specific systems and a sevenfold profession level spread in net image effect for proprietary models, from $+5.7$ points on Physician questions to $+39.8$ points on Public Health Nurse questions. As a limitation, JMed48k evaluates licensing examination performance rather than clinical deployment, and the audit measures observable answer changes rather than human like reasoning. We release the corpus, image taxonomy, evaluation subset, and paired audit protocol for reproducible, profession stratified evaluation of medical vision language models.

\section*{Acknowledgments.}
This paper is based on results obtained from a project, JPNP25006, commissioned by the New Energy and Industrial Technology Development Organization (NEDO). The computation was carried out using the computer resource offered under the category of General Projects (Project ID: p260010396) by Research Institute for Information Technology, Kyushu University.

\bibliographystyle{abbrvnat}
\bibliography{refs}

\newpage
%======================================================================
\appendix
%======================================================================
% Use natural-bottom layout in the appendix so paragraph breaks don't get
% stretched to fill short pages (NeurIPS' \flushbottom is mandatory for
% the 9-page main body but appendices can flow naturally).
\raggedbottom

\section{Dataset Construction and Quality Control}\label{app:curation}

This appendix documents the construction of JMed48k from official MHLW examination PDFs to the final scored evaluation subset. It covers the source materials, the PDF-to-JSON extraction pipeline, the quality-control procedures, the correction categories applied during validation, the gold-answer alignment, the items excluded from scoring, the per-profession composition of the released subsets, and the released JSON schema. Scoring rules and answer normalisation are described separately in Appendix~\ref{app:scoring}, and image taxonomy details in Appendix~\ref{app:image_taxonomy}.

\subsection{Source materials and PDF-to-JSON extraction pipeline}

For each (profession, year) cell, we construct structured records from official examination materials released by the Japanese Ministry of Health, Labour and Welfare (MHLW). The source materials consist of three document types: a \emph{question booklet} containing question stems and answer-choice tables, a \emph{figure booklet} containing the medical images referenced by the questions, and an \emph{answer-key document} listing the official correct answers. The three documents are processed separately and joined by \texttt{(profession, year, section, question\_number)}.

\paragraph{Layout extraction.} Each booklet PDF is rasterised at high resolution and processed with MinerU~\citep{wang2024mineru}, a layout-aware document content extraction system. We invoke it with the Japanese OCR model and disable table detection, which empirically misclassifies medical-image captions as table content. MinerU produces a sequential markdown stream that interleaves extracted text, formulas, and image regions, together with an index of bounding boxes, page numbers, and reading order.

\paragraph{Image--caption pairing.} MHLW figure booklets reference each medical image with a structured caption of the form ``No.\,$X$ ($Y$ Problem $Z$)'' where $X$ is the figure-booklet image index, $Y$ is the question section letter, and $Z$ is the question number within the section. A finite-state-machine parser scans the markdown stream and pairs each extracted image with its caption by matching this pattern, producing per-question image files named \texttt{\{specialty\}\_\{year\}\_\{section\}\_Q\{number\}\_\{role\}.png}, where \texttt{role} distinguishes content images, answer-choice images (used by image-as-options items), and supplementary images.

\paragraph{Vertical-orientation handling.} Roughly a quarter of MHLW figure booklets are typeset in vertical (portrait) orientation. In this layout, the OCR pipeline routinely captures vertical caption text together with adjacent image content and produces malformed image--caption pairs. We therefore process each figure booklet twice, once in its original orientation and once in a $90^{\circ}$ rotated copy, and merge the extractions per image, retaining the orientation that yielded a cleaner caption.

\paragraph{Answer-key joining.} The answer-key document is parsed independently and the gold answer is written into the corresponding question record's \texttt{correct\_answer} field. Gold answers are never produced by the extractor or by any model. The pipeline outputs one structured JSON record per question; the released schema and example records are documented in Section~\ref{app:curation_schema}.

\subsection{Quality control and correction log}\label{app:curation_qc}

Every extracted record passes a battery of automated consistency checks before inclusion in the release. The checks detect missing required fields, malformed option keys, duplicate question identifiers within a section, gold answers outside the available option set, unresolved image references, and inconsistencies between the modality-routing flag and the presence of image references. Records flagged by these checks are reviewed by a human annotator against the original PDF materials.

To reduce OCR and formatting noise further, we engaged a professional Japanese-language data-correction vendor to review the PDF-to-JSON pipeline output and the extracted image files. The vendor review focused on Japanese text normalisation, malformed symbols, chemical and isotope notation, option ordering, image-path consistency, and image--question pairing. Gold answers were cross-checked against the official MHLW answer-key documents. The vendor did not generate new questions, rewrite clinical content, or relabel the image taxonomy; image taxonomy annotation is documented separately in Appendix~\ref{app:image_taxonomy}.

Across the corpus more than $26{,}700$ corrections were applied during validation. Table~\ref{tab:curation_categories} reports the per-category breakdown.

\begin{table}[t]
  \centering
  \caption{\textbf{Correction categories applied during validation.} Most edits are mechanical normalisations: image filename and option-key ordering together account for $\approx 92\%$ of the log. Smaller categories involving image--question pairing or gold-answer alignment were resolved only by checking the source PDFs and the official MHLW answer-key documents.}
  \label{tab:curation_categories}
  \footnotesize
  \begin{tabular}{l r l}
    \toprule
    \rowcolor{HeaderBand}
    Category & Count & Main issue fixed \\
    \midrule
    Image filename normalisation     & $12{,}120$ & Path/prefix unification across years \\
    Option-key ordering              & $12{,}468$ & Canonical alphabetic / numeric / symbolic label ordering \\
    Modality flag (\texttt{text\_only}) & $624$   & Reconciliation between routing flag and image references \\
    Image-field unification          & $601$    & Consistency between \texttt{content\_img} and \texttt{answer\_img} \\
    Multi-answer formatting          & $513+35$ & Cleanup of string, flat-list, and nested-list answer encodings \\
    Gold-answer cross-checks         & $28+27$  & PDF-verified corrections against official answer materials \\
    Other                            & ${\sim}300$ & Misc.\ schema-consistency edits \\
    \midrule
    \textbf{Total}                   & $\mathbf{26{,}700+}$ & \\
    \bottomrule
  \end{tabular}
\end{table}

\paragraph{Interpretation of the correction log.} The total correction count is large but is driven primarily by mechanical normalisation: image filename normalisation and option-key ordering together account for approximately $92\%$ of all edits, and these standardise file paths, label conventions, and schema fields without rewriting clinical content. A smaller number of corrections involve image--question pairing or gold-answer alignment; for these scoring-relevant categories, an edit was applied only when the source PDF or the official MHLW answer-key document provided direct evidence. No correction introduced model-generated content, model-generated answers, or unsupported relabelling.

\subsection{Gold-answer alignment and excluded items}\label{app:curation_gold}

Gold answers are aligned from the official MHLW answer-key documents rather than generated by any extractor or any language model. For each (profession, year) cell, the official answer key is parsed by section and question number, and the resulting answer is written into the corresponding record's \texttt{correct\_answer} field. During validation, gold answers were cross-checked against the original answer-key PDFs.

After alignment, four legitimate gold-answer structures remain in the corpus, and all are honoured by the scoring code (Appendix~\ref{app:scoring}). \emph{Single-key} answers store one option key and are the dominant case across all professions. \emph{Multi-key sets} store more than one option key when the examination requires multiple correct choices; this format appears with non-trivial frequency in the Dentist and Pharmacist examinations. \emph{Alternative-accepting} answers store a list of acceptable answer sets ($95$ items in the scored Eval) for items where the official material recognises more than one acceptable answer; the scorer accepts a prediction that matches any one of the alternatives. \emph{Numeric-slot} answers store one or more digit-string slots for calculation items, used predominantly in the Pharmacist, Physical Therapist, and Occupational Therapist examinations.

\paragraph{The $93$ excluded items.} The scored subset on which all reported accuracies are computed contains $12{,}484$ questions, after dropping $93$ items from the five-year evaluation split for which MHLW did not publish a single reproducible official target. These $93$ items fall into three operational categories: (a) items that MHLW retroactively \emph{withdrew} after the examination; (b) items for which MHLW issued a \emph{multiple-acceptable-answer} ruling without committing to a canonical key; and (c) items whose published answer remained \emph{unresolved} for scoring, for example because official errata did not converge to a single usable target. These items are retained in the released corpus with \texttt{correct\_answer = null} for transparency, but they are excluded from all reported accuracy denominators. All main evaluation tables are computed on the same $12{,}484$ scored questions: $9{,}905$ text-only and $2{,}579$ with images.

\subsection{Profession-level composition}

Table~\ref{tab:dataset_profession_counts} reports the per-profession composition of JMed48k. The full-corpus columns summarise all official questions collected from MHLW examination materials across 2005--2025 and the $9{,}646$ questions that reference one or more medical images. The scored JMed48k-Eval columns report the denominator used in all main evaluation tables, after restricting to the most recent five available years per profession (2021--2025 for eight professions; 2020--2024 for Public Health Nurse, Midwife, and Nurse) and excluding the $93$ items without a usable official gold answer. The scored subset contains $12{,}484$ questions: $9{,}905$ text-only and $2{,}579$ with images.

\input{tables/T_dataset_profession_counts}

\subsection{Released JSON schema and example records}\label{app:curation_schema}

The released corpus is stored as JSON records, with one record corresponding to one examination question. Each per-cell file contains an array of question records under the key \texttt{questions}; the cell-level keys \texttt{specialty} and \texttt{year} are determined by the directory layout (\texttt{JMed48k/\{specialty\}/\{specialty\}\_\{year\}/\{year\}\_CORRECTED.json}). Field names are aligned with how each field is consumed downstream: \texttt{text\_only} is a modality-routing flag, not a difficulty label, and the per-question image-type aggregation and \texttt{image\_as\_options} flag (carried in companion taxonomy annotations; Appendix~\ref{app:image_taxonomy}) are layout/category metadata, not content judgements. The fields are summarised in Table~\ref{tab:schema_fields}.

\begin{table}[t]
  \centering
  \caption{\textbf{Major fields of a JMed48k JSON record} (one record per examination question). \texttt{specialty} and \texttt{year} are encoded in the directory layout rather than in the record itself.}
  \label{tab:schema_fields}
  \footnotesize
  \setlength{\tabcolsep}{4pt}
  \begin{tabularx}{\linewidth}{l l X X}
    \toprule
    \rowcolor{HeaderBand}
    Field & Type & Meaning & Downstream use \\
    \midrule
    \texttt{section} & string & Examination section letter & Record identity, PDF joining \\
    \texttt{question\_number} & integer & Question number within the section & Record identity, PDF joining \\
    \texttt{question\_text} & string & Japanese question stem & Prompt input \\
    \texttt{options} & object & Choices keyed by option label & Prompt input and scoring \\
    \texttt{correct\_answer} & list / object / null & Official answer from the MHLW answer key & Scoring eligibility and scoring \\
    \texttt{text\_only} & boolean & Whether the released item has any image references & Modality routing \\
    \texttt{img.content\_img} & string / list & Image references attached to the question stem & Multimodal input \\
    \texttt{img.answer\_img} & string / list & Image references attached to the answer choices & Image-as-options input \\
    \texttt{text\_reference} & string / null & Optional cross-reference to a paired earlier question & Multi-question links \\
    \bottomrule
  \end{tabularx}
\end{table}

The companion image-taxonomy annotations (Appendix~\ref{app:image_taxonomy}) carry the per-question dominant primary image type, the set of image types attached to the question, and the \texttt{image\_as\_options} flag indicating whether the visual content carries the answer choices themselves. The latter is the basis for the positive-control audit in Appendix~\ref{app:image_options}.

The taxonomy assigns each image element one of nine labels; eight of these are substantive primary types and \emph{Other/Unclear} is a fallback category. The Other/Unclear bucket contains image references for which MHLW does not publicly redistribute the visual content (most commonly for privacy or rights reasons such as patient portraits), so the taxonomy classifier has no image to inspect; those records still preserve the existence of an image reference. Other/Unclear accounts for $44$ of $20{,}142$ image-level annotations ($0.2\%$); we retain it in the released annotations for transparency but exclude it from type-stratified analyses.

\paragraph{Example: text-only single-label question (Physician 2024 \S B Q1).}
\begin{CJK}{UTF8}{min}
\noindent\fbox{\parbox{0.97\linewidth}{\footnotesize\ttfamily
\{\\
\hspace*{1em}"question\_number": 1,\\
\hspace*{1em}"section": "B",\\
\hspace*{1em}"question\_text": "医師の指示があっても採血できない職種はどれか。",\\
\hspace*{1em}"options": \{\\
\hspace*{2em}"a": "看護師",\\
\hspace*{2em}"b": "助産師",\\
\hspace*{2em}"c": "保健師",\\
\hspace*{2em}"d": "薬剤師",\\
\hspace*{2em}"e": "臨床検査技師"\\
\hspace*{1em}\},\\
\hspace*{1em}"correct\_answer": ["D"],\\
\hspace*{1em}"text\_only": true,\\
\hspace*{1em}"img": \{"content\_img": "", "answer\_img": ""\},\\
\hspace*{1em}"text\_reference": null\\
\}
}}
\end{CJK}

\paragraph{Example: with-images single-label question (Physician 2024 \S A Q38).}
\begin{CJK}{UTF8}{min}
\noindent\fbox{\parbox{0.97\linewidth}{\footnotesize\ttfamily
\{\\
\hspace*{1em}"question\_number": 38,\\
\hspace*{1em}"section": "A",\\
\hspace*{1em}"question\_text": "43歳の女性。右眼が赤いことに気付き来院した。外傷や手術の既往はなく自覚症状はない。眼脂を認めない。右眼の写真(別冊No.14)を別に示す。最も考えられるのはどれか。",\\
\hspace*{1em}"options": \{\\
\hspace*{2em}"a": "霰粒腫",\\
\hspace*{2em}"b": "睫毛乱生",\\
\hspace*{2em}"c": "眼瞼外反",\\
\hspace*{2em}"d": "結膜下出血",\\
\hspace*{2em}"e": "流行性角結膜炎"\\
\hspace*{1em}\},\\
\hspace*{1em}"correct\_answer": ["D"],\\
\hspace*{1em}"text\_only": false,\\
\hspace*{1em}"img": \{"content\_img": "2024\_images/A38\_content.png", "answer\_img": ""\},\\
\hspace*{1em}"text\_reference": ""\\
\}
}}
\end{CJK}

The companion taxonomy file annotates the \S A Q38 image as \texttt{Clinical Photograph} with \texttt{image\_as\_options} = false, since the visual content supports the question stem rather than carrying the answer choices.

\section{Scoring, Answer Normalization, and Denominator Handling}\label{app:scoring}

This appendix defines the scoring contract used for every accuracy reported in the paper. Model predictions are returned as free-form strings or model-specific JSON wrappers, while JMed48k items inherit several answer structures from the official MHLW examination materials. We therefore apply a deterministic normalisation pipeline that reduces model outputs to canonical option keys or numeric slot values before comparing them with the official gold answers. The pipeline handles conventions common in Japanese medical examinations, including full-width punctuation, circled-digit option labels, and natural-language answer phrases. We also define how refusals, parse failures, inference errors, and missing outputs are counted, so that all reported accuracies use fixed and comparable denominators.

\subsection{Scored item universe and evaluation contract}

All reported accuracies are computed on the scored JMed48k-Eval subset defined in Appendix~\ref{app:curation}. After excluding items without a usable official MHLW gold answer, the scored subset contains $12{,}484$ questions: $9{,}905$ text-only and $2{,}579$ with images. Text-only accuracy is computed on the $9{,}905$ text-only items, with-images accuracy on the $2{,}579$ image-bearing items, and the paired image-removal audit on the same $2{,}579$ image-bearing items evaluated twice, once with the linked images and once after removing all visual content.

The scorer receives two inputs: a model output and the corresponding item record. The item record provides the official gold answer, the item-specific option set, and the answer-format type implied by the structure of \texttt{correct\_answer}. The scorer first normalises the prediction into canonical option keys or slot values, then compares the normalised prediction against the official gold answer. Non-answer outputs, parse failures, inference errors, and missing outputs are retained in the denominator and scored as incorrect, as described in Section~\ref{app:scoring_failure}.

\subsection{Gold-answer formats inherited from MHLW examinations}\label{app:scoring_formats}

JMed48k preserves the answer structures used by the official MHLW examination materials. These structures are recorded in the \texttt{correct\_answer} field of each released JSON record (Appendix~\ref{app:curation_schema}); they are not generated by models and are not introduced as benchmark-specific labels. Across the $12{,}484$ scored items, the field takes one of four distinct structures, summarised in Table~\ref{tab:scoring_gold_formats}. Two further behaviours --- \emph{numeric-slot scoring} and \emph{ordering scoring} --- are detector-triggered overlays on top of these structures and are described at the end of this subsection.

\begin{table}[t]
  \centering
  \caption{\textbf{Gold-answer structures in JMed48k-Eval scored items.} Counts are over the $12{,}484$ scored questions and total exactly $12{,}484$. ``Single-key'' and ``multi-key'' refer to the cardinality of the official answer set; the alternative-accepting form ($\mathit{list[list[str]]}$) allows the official material to list more than one acceptable answer set, where each acceptable set is itself either a single key or a multi-key combination. Three sub-patterns of the alternative-accepting form are described in the prose below.}
  \label{tab:scoring_gold_formats}
  \footnotesize
  \setlength{\tabcolsep}{4pt}
  \begin{tabularx}{\linewidth}{l l r X}
    \toprule
    \rowcolor{HeaderBand}
    Structure & Example & Count & Scoring rule \\
    \midrule
    Single key                & \texttt{["3"]}             & $9{,}673$ & Normalised prediction equals the gold key \\
    Multi-key set             & \texttt{["2","4"]}         & $2{,}716$ & Normalised prediction set equals the gold set \\
    Alternative-accepting set & \texttt{[["1"],["4"]]}     & $95$      & Accept if the normalised prediction matches any one of the alternatives \\
    \bottomrule
  \end{tabularx}
\end{table}

\emph{Single key.} The official answer is a single option key, such as \texttt{"3"}. This is the dominant case across all 11 professions. Scoring requires the normalised prediction to equal the gold key under the rules of Section~\ref{app:scoring_rules}.

\emph{Multi-key set.} The official answer requires more than one option, recorded as a list of option keys such as \texttt{["2","4"]}, \texttt{["A","B","E"]}. Scoring requires the normalised prediction \emph{set} to equal the gold set; partial matches receive no credit. Multi-key items appear in all professions, with the highest density in the Dentist and Pharmacist examinations.

\emph{Alternative-accepting set.} For some MHLW items, the official answer material lists \emph{more than one} acceptable answer set; any one of these alternatives is treated as correct. The released \texttt{correct\_answer} encodes this as a list of lists, in which each inner list is itself a complete acceptable answer (a single key or a multi-key combination). The scorer accepts a prediction that exactly matches any one of the alternatives, and rejects everything else; partial credit is not awarded. Three sub-patterns occur in JMed48k-Eval: (i)~\emph{single-key alternatives}, where each inner list is one option, e.g.\ \texttt{[["1"],["4"]]} (``\,answer either $1$ or $4$\,''); (ii)~\emph{multi-key alternatives}, where each inner list is a fixed-size combination, e.g.\ \texttt{[["2","4"],["3","4"]]} (``\,answer either the pair $\{2,4\}$ or the pair $\{3,4\}$\,''); (iii)~\emph{mixed-cardinality alternatives}, where inner lists have different lengths, e.g.\ \texttt{[["A","B","D","E"],["B","D","E"]]} (``\,answer either the four-key set or the three-key set\,''). The alternative-accepting form is uncommon overall (95 items, $0.8\%$ of the scored Eval) but contains items where MHLW officially recognises that more than one answer set is acceptable, and the scorer therefore must not collapse it into a single canonical set.

\paragraph{Numeric-slot scoring.} For calculation-style items, the gold answer is a list of digit-string slots, such as \texttt{["1","2"]}, with the question text indicating the unit and the slot order. The scorer normalises both the gold answer and the prediction by joining their digits in order and stripping non-digit characters; if the joined digit strings agree, the item is correct. Numeric-slot recognition triggers when (i)~every component of the gold answer is a digit string of the form \texttt{\textbackslash d+(.\textbackslash d+)?}, \emph{and} (ii)~the question text contains a numeric-entry cue such as ``\begin{CJK}{UTF8}{min}求めよ\end{CJK}'' (compute), ``\begin{CJK}{UTF8}{min}四捨五入\end{CJK}'' (round), ``\begin{CJK}{UTF8}{min}小数点\end{CJK}'' (decimal point), or ``\begin{CJK}{UTF8}{min}解答:\end{CJK}'' (answer:). Numeric-slot items appear primarily in the Pharmacist, Physical Therapist, and Occupational Therapist examinations. The numeric-slot detector is orthogonal to the structural format above: a numeric-slot item is itself recorded as a single-key, multi-key, or alternative-accepting structure, with the slot values stored as digit strings inside that structure.

\paragraph{Ordering scoring.} When the question requires a sequence rather than an unordered set, the scorer compares the prediction to the gold sequence position by position rather than after sorting. Ordering recognition triggers on a multi-element gold answer combined with a sequence cue in the question text, such as ``\begin{CJK}{UTF8}{min}並べよ\end{CJK}'' (arrange), ``\begin{CJK}{UTF8}{min}順番に\end{CJK}'' (in order), or an explicit circled-digit arrow chain (\textcircled{1}, \textcircled{2}, \textcircled{3}) connected by arrows in the question text. As with numeric-slot scoring, the ordering detector is orthogonal to the structural format: an ordering item can be encoded as a multi-key set or as alternative-accepting alternatives where each alternative is itself a sequence.

\subsection{Normalisation principles}

The normaliser is deterministic and conservative. A rule succeeds only if it returns an item-specific valid option key derived from the record's \texttt{options} field, either directly or via a reverse lookup over the option text. Candidate labels not in the valid option set for that item are rejected. The rules are applied in a fixed order, and the first rule that yields a valid key wins.

The normaliser is designed to recover standard answer forms produced by instruction-following models, not to grade explanations semantically. It accepts exact option labels, full option text, common answer-keyword phrases, boxed labels, parenthesised labels, and circled-digit labels. \emph{It does not use an LLM judge, and does not infer correctness from medical-reasoning text.}

\subsection{Answer normalisation rules}\label{app:scoring_rules}

The normaliser is implemented in \texttt{compare\_answers} of the released evaluation script. It applies the following rules in order; the first rule that yields a valid option key wins.

\paragraph{Preprocessing.}
\begin{itemize}\setlength{\itemsep}{1pt}\setlength{\topsep}{1pt}
\item \textbf{Rule 0 (cleanup).} Strip ASCII control characters in the range \texttt{U+0000}--\texttt{U+0008}, \texttt{U+000B}--\texttt{U+000C}, \texttt{U+000E}--\texttt{U+001F}, and \texttt{U+007F} (whitespace characters tab, line feed, and carriage return are preserved), then collapse internal whitespace. For label-format items, uppercase alphabetic labels. Control characters appear when model outputs are serialised through JSON wrappers, when backslashes in LaTeX-style outputs are decoded as escape sequences, or when provider-specific inference stacks emit malformed text fragments.
\end{itemize}

\paragraph{Exact text matching.}
\begin{itemize}\setlength{\itemsep}{1pt}\setlength{\topsep}{1pt}
\item \textbf{Rule 1 (full option-text reverse lookup).} If the cleaned prediction matches the full text of one of the option choices, ignoring case and harmless surrounding whitespace, return that option's key. This rule is applied before label-prefix extraction so that full option-text answers are credited and short-label heuristics do not accidentally match label-like characters embedded in longer option strings.
\end{itemize}

\paragraph{Structural label extraction.}
\begin{itemize}\setlength{\itemsep}{1pt}\setlength{\topsep}{1pt}
\item \textbf{Rule 2 (label-prefix pattern).} If the prediction starts with a valid label followed by an ASCII period, a full-width period (\texttt{U+FF0E}), or an ideographic period (\texttt{U+3002}), and then additional text, return the leading label. This handles outputs such as \texttt{"3. Tryptophan"}.
\item \textbf{Rule 3 (bare label).} If the prediction, after stripping trailing punctuation, is exactly a valid label, return it. This handles outputs such as \texttt{"a"}, \texttt{"A."}, or \texttt{"3"}.
\item \textbf{Rule 4 (boxed notation).} If the prediction contains \verb|\boxed{N}| where $N$ is a valid label, return $N$. This handles LaTeX-style outputs from reasoning-tuned models. The rule also matches the corrupted form \verb|oxed{N}| that can appear when a leading backspace control character is stripped from \verb|\boxed{N}| during Rule~0 cleanup.
\item \textbf{Rule 5 (trailing parenthesised label).} If the prediction ends with a label inside ASCII parentheses \texttt{(N)} or full-width parentheses (\texttt{U+FF08}, \texttt{U+FF09}), and $N$ is a valid label, return $N$. This handles outputs such as \texttt{"The answer is (3)"}.
\end{itemize}

\paragraph{Language-specific answer patterns.}
\begin{itemize}\setlength{\itemsep}{1pt}\setlength{\topsep}{1pt}
\item \textbf{Rule 6 (answer-keyword patterns).} Scan the prediction for natural-language ``answer is $N$''-style phrases. The implemented patterns are case-insensitive English (\texttt{answer is/:}, optionally prefixed by \texttt{the}/\texttt{correct}/\texttt{final}) and four Japanese forms ``\begin{CJK}{UTF8}{min}正解は\end{CJK}\,$N$'', ``\begin{CJK}{UTF8}{min}解答は\end{CJK}\,$N$'', ``\begin{CJK}{UTF8}{min}答えは\end{CJK}\,$N$'', ``\begin{CJK}{UTF8}{min}最終的な回答は\end{CJK}\,$N$'', plus the explicit choice phrase ``\begin{CJK}{UTF8}{min}選択肢\end{CJK}\,$N$\,\begin{CJK}{UTF8}{min}が正しい\end{CJK}/\begin{CJK}{UTF8}{min}が正解\end{CJK}/\begin{CJK}{UTF8}{min}を選ぶ\end{CJK}/\begin{CJK}{UTF8}{min}を選択\end{CJK}''. If a pattern matches and $N$ is a valid label, or if $N$ is a full option text recoverable through reverse lookup, return the corresponding key.
\item \textbf{Rule 7 (trailing circled-digit label).} If the prediction ends with a circled digit \textcircled{1}, \textcircled{2}, \ldots, \textcircled{10} and the digit maps to a valid option key for the item, return that key. This handles outputs that mimic the official Japanese examination typography.
\end{itemize}

\paragraph{Fallback.} If none of Rules~1--7 yields a valid key, the cleaned string from Rule~0 is compared as-is to the gold answer (with both sides sorted as a multi-element set match); mismatches are scored as incorrect.

\subsection{Scoring rules by answer format}

After normalisation, scoring is exact and format-specific. \emph{No partial credit is awarded under any format.}
\begin{itemize}\setlength{\itemsep}{1pt}\setlength{\topsep}{1pt}
\item \textbf{Single key.} The normalised prediction must equal the official gold key.
\item \textbf{Multi-key set.} The set of normalised predicted keys must equal the gold set. Predictions encoded as a JSON array \texttt{["2","4"]} are consumed directly; predictions encoded as a single arrow-separated string \texttt{["B->E->C"]} (using ASCII arrows or full-width arrows) are split into the array \texttt{["B","E","C"]} before normalisation. Comma-separated, conjunction-joined, or sentence-form multi-answer predictions are not split by the scorer; the JSON-output schema enforced during inference is the primary mechanism for emitting multi-answer predictions.
\item \textbf{Alternative-accepting.} The scorer evaluates the normalised prediction against each alternative in the official list and accepts the item if any one alternative matches under the rule appropriate for that alternative's structure (single-key equality or multi-key set equality, including position-by-position equality when the alternative is itself an ordering sequence).
\item \textbf{Ordering.} The normalised predicted sequence is compared to the gold sequence position by position, without sorting.
\item \textbf{Numeric slot.} The gold answer and the prediction are each joined into a single digit string (for example \texttt{["9","0"]}\,$\rightarrow$\,\texttt{"90"}; non-digit characters such as decimals, percent signs, or units are stripped). The item is correct iff the joined digit strings are equal. The scorer does not perform value-preserving decimal normalisation: \texttt{"9.0"} and \texttt{"9"} would join to different digit strings (\texttt{"90"} vs.\ \texttt{"9"}). MHLW numeric-entry items are released as integer-slot answers, so this matters only if a model's output mixes decimal forms with the slot convention.
\end{itemize}

\subsection{Refusals, parse failures, inference errors, and missing outputs}\label{app:scoring_failure}

We treat non-answer responses uniformly as incorrect predictions. This conservative policy keeps denominators fixed across models and prevents systems from improving apparent accuracy by abstaining on difficult items. A licensing-exam benchmark measures answer correctness rather than selective abstention; therefore a model that refuses, emits an unparseable answer, or fails to return a recoverable output has not produced a correct answer to the item. The following response types are retained in the denominator and scored as incorrect:

\begin{itemize}\setlength{\itemsep}{1pt}\setlength{\topsep}{1pt}
\item \textbf{Calibrated refusal.} A model output that declines to answer or states that the necessary information is unavailable. In our suite, calibrated refusals appear most prominently for the GPT-5 family under image removal, where a small fraction of responses return a well-formed JSON object with an empty \texttt{answer} array and a Japanese-language explanation that the figure is not provided. These items are scored as incorrect in the main accuracy and paired image-removal tables. Whether such refusals are calibrated to genuine uncertainty rather than safety over-conservatism is examined separately as an auxiliary forced-guess control in Appendix~\ref{app:refusal}; that auxiliary analysis does not change the headline scoring rule.
\item \textbf{Parse failure.} A response that cannot be normalised to a valid option key or required numeric slot value after Rules~0--7. The cleaned-string fallback then yields a non-key string that does not match the gold answer, and the item is scored as incorrect.
\item \textbf{Inference-stack error.} An item for which inference returned an HTTP error, a provider-side timeout, a vLLM stack failure, a malformed JSON wrapper that bypassed schema validation, or any other unrecoverable execution error. The scorer writes \texttt{predicted=null}, \texttt{correct=false}, and an \texttt{error} field describing the failure.
\item \textbf{Missing item.} An item for which no result is recoverable from the per-cell run or its backups. Such items are treated as incorrect by construction so that table cells share a constant denominator.
\end{itemize}

The shared scored denominator is $12{,}484$ items overall: $9{,}905$ items for the text-only table and $2{,}579$ items for the with-images table. The paired image-removal audit uses the same $2{,}579$ image-bearing items evaluated under two input conditions, and the four answer-transition states $p_{11}, p_{10}, p_{01}, p_{00}$ are reported on this paired denominator.

\section{Image Taxonomy Annotation}\label{app:image_taxonomy}

This appendix documents the image taxonomy used in JMed48k, the annotation units on which labels are assigned, and the model-assisted annotation pipeline that produced the released labels. The taxonomy contains \emph{eight primary inspectable visual types} and an auxiliary \emph{Other/Unclear} bucket. Labels are assigned at the visual-element level, aggregated to source PNGs and questions, and used for the image-type-stratified analyses in Section~\ref{sec:image-removal} and Appendix~\ref{app:image-type-matrix}.

The taxonomy is a \emph{visual-content} taxonomy rather than a clinical-modality ontology. For example, X-ray, CT, MRI, ultrasound, angiography, and nuclear-medicine images are grouped under Radiology/Scan. This coarser grouping is designed to support robust analysis across 11 healthcare professions and heterogeneous official examination formats.

\subsection{Scope and annotation units}

A single extracted source PNG can carry one or more visual elements. In simple cases, the PNG contains a single radiograph, clinical photograph, chart, or diagram. In multi-element cases, the PNG may contain a multi-panel layout (e.g.\ several CT slices from different time points; pathology micrographs at different magnifications) or an image-as-options layout where each panel \emph{is} one of the answer choices. We therefore annotate the visual element, rather than the raw PNG file, as the primary unit. The released metadata preserves this distinction at three levels:

\begin{itemize}\setlength{\itemsep}{1pt}\setlength{\topsep}{1pt}
\item \textbf{Image-level (visual element).} One primary type per visual element produced by sub-image segmentation. This is the unit of taxonomy annotation.
\item \textbf{Source PNG.} The extracted file path plus the file's dominant primary type, computed from its element-level labels.
\item \textbf{Question.} The set of all primary types attached to any visual element referenced by the question (released as the per-question \texttt{image\_types} field), together with a single dominant primary type used for analyses that require one type per question.
\end{itemize}

This three-level distinction is why the full corpus contains $20{,}142$ image-level annotations across $12{,}972$ source PNG files, with $9{,}646$ image-bearing questions.

\subsection{Taxonomy definitions and borderline decisions}

Table~\ref{tab:image_taxonomy_definitions} defines the eight primary inspectable visual types and the auxiliary Other/Unclear bucket. Definitions are operational and were written to guide human-annotator judgement when resolving machine-assigned labels.

\begin{table*}[t]
  \centering
  \caption{\textbf{Visual-content taxonomy used in JMed48k.} The first eight rows are the primary inspectable visual types used in image-type-stratified analyses. \emph{Other/Unclear} is an auxiliary bucket retained in released metadata but excluded from type-stratified performance analyses.}
  \label{tab:image_taxonomy_definitions}
  \footnotesize
  \begin{tabularx}{\textwidth}{p{2.4cm} X X}
    \toprule
    \rowcolor{HeaderBand}
    Type & Definition & Common borderline decisions \\
    \midrule
    Clinical Photograph & Visible-light photograph of a body region, lesion, surgical field, gross specimen, or external clinical finding. & Endoscopic and laparoscopic internal views go to \emph{Endoscopy}. Gross specimen photographs are \emph{Clinical Photograph}; only microscopic views are \emph{Pathology/Microscopy}. \\
    Radiology/Scan & X-ray, CT, MRI, ultrasound, angiography, nuclear-medicine image, or reconstructed scan view. & Macroscopic radiographs of pathology specimens remain \emph{Radiology/Scan}. \\
    Pathology/Microscopy & Histology, cytology, electron microscopy, or any optical/electron microscopic view. & Gross specimens are \emph{Clinical Photograph}; drawn microscopic structures are \emph{Diagram/Schematic}. \\
    Endoscopy & Endoscopic, bronchoscopic, laparoscopic, colonoscopic, or other scope-acquired internal views. & External photographs of the body surface or scope device are \emph{Clinical Photograph}. \\
    Chart/Graph/Table & Quantitative plots, lab-result tables, statistical summaries, plotted curves, or data tables. & A single physiological trace presented as the primary object is \emph{Waveform/Signal}; schematic workflows are \emph{Diagram/Schematic}. \\
    Diagram/Schematic & Drawn anatomy, mechanisms, procedures, pathways, workflows, or conceptual schematics; not plots of data. & Small-molecule structural formulas are \emph{Chemical/Molecular} even when stylised. Drawn microscopic structures are \emph{Diagram/Schematic}. \\
    Chemical/Molecular & 2D structural formulas, line-angle molecular drawings, simple reaction schemes, and chemical structures. & Large cellular pathway diagrams are \emph{Diagram/Schematic} unless the primary task is molecular-structure recognition. \\
    Waveform/Signal & Physiological time-series traces such as ECG, EEG, respiratory, or pressure strips presented as the primary content. & A trace embedded as one of several plotted curves on common axes is \emph{Chart/Graph/Table}. \\
    \midrule
    Other/Unclear & \emph{Auxiliary bucket.} Image references whose visual content MHLW does not publicly redistribute (most commonly for privacy or rights reasons such as patient portraits), so the classifier has no inspectable image; also visual content that does not fit any of the eight primary types. & Retained for transparency. Excluded from type-stratified performance analyses. \\
    \bottomrule
  \end{tabularx}
\end{table*}

\subsection{Annotation pipeline and human annotator resolution}

The annotation pipeline is model-assisted with human annotator verification. We do not perform full independent double annotation over the entire corpus. Instead, a first classifier assigns provisional labels, a second classifier is invoked selectively on flagged items, and a human annotator resolves disagreements and borderline cases.

\paragraph{First pass.} Each source PNG is sent to GPT-4.1 via the Azure OpenAI API with a system prompt that defines the visual-content labels and asks the model to (i) identify visual elements within multi-panel images, (ii) assign one primary type per element, and (iii) flag image-as-options layouts. Temperature is fixed to $0$. The first-pass prompt used an earlier taxonomy in which endoscopic views were not yet separated as their own primary type.

\paragraph{Flagged second pass.} Items labelled \emph{Other/Unclear} or otherwise flagged as ambiguous are reclassified with Gemini~2.5 Pro using an updated prompt that introduces \emph{Endoscopy} as a separate primary type. The second pass repairs a known taxonomy-evolution issue: endoscopic and laparoscopic views were previously distributed between \emph{Clinical Photograph} and \emph{Other/Unclear} in the first-pass labels, and the second pass reassigns them to \emph{Endoscopy}. The second classifier is not run on every item, so the pipeline relies on targeted review rather than independent dual annotation.

\paragraph{Human-annotator resolution.} A human annotator reviews items where classifiers disagree or where an item is flagged for manual review. The human annotator applies the operational definitions and borderline rules in Table~\ref{tab:image_taxonomy_definitions} and writes the final released label. Human-annotator decisions are logged together with the source-file identifier and final label so that the released annotation set is reproducible.

\subsection{Aggregation to question-level analysis labels}

Downstream analyses use two question-level representations of the image-level labels.

The first is a \emph{multi-label set}, stored in the per-question \texttt{image\_types} field, containing every primary type attached to any visual element referenced by the question. This representation is used wherever the analysis allows a question to contribute to multiple type memberships, including the \emph{Eval Q (any)} column of Table~\ref{tab:dataset_image_types}.

The second is a \emph{single dominant primary type}, used for one-type-per-question stratification in the per-(model, image type) heatmap of Appendix~\ref{app:image-type-matrix} and the cohort-mean tables in Section~\ref{sec:image-removal}. The dominant type is computed deterministically from the question's image-level labels by the released annotation script; the multi-label set is preserved in the released metadata, so any analysis can recover both the dominant view and the multi-type membership view.

\subsection{Image-type distribution}

Table~\ref{tab:dataset_image_types} reports the taxonomy distribution at three granularities: image-level annotations over the full corpus, source-PNG counts under dominant-type assignment, and question-level counts in JMed48k-Eval. Image-level counts exceed source-PNG counts because a single source PNG may contain multiple visual elements. Eval Q (any) can sum to more than the with-images denominator because a question can reference multiple visual types; Eval Q (dom.) assigns each scored with-images question to one dominant type and sums to $2{,}579$, the with-images denominator used throughout the paper.

\input{tables/T_dataset_image_types}

\section{Per-profession breakdown}\label{app:profession_breakdown}

Table~\ref{tab:profession_breakdown} reports the profession-level composition of the scored JMed48k-Eval subset. It complements the corpus-level construction summary in Appendix~\ref{app:curation} and the image-taxonomy distribution in Appendix~\ref{app:image_taxonomy}. For each profession, we report the number of scored questions with images, their dominant primary image types, the number of image-as-options (IO) items, and the answer-format distribution used by the scorer.

This table serves two purposes. First, it shows that visual content is not uniformly distributed across professions: some professions are dominated by Radiology/Scan or Clinical Photograph, whereas others contain mostly Chart/Graph/Table, Diagram/Schematic, or Chemical/Molecular content. Pharmacist is dominated by Chart/Graph/Table and Chemical/Molecular; Public Health Nurse, Midwife, and Nurse are dominated by Chart/Graph/Table and Diagram/Schematic; Physician and Radiological Technologist carry the largest Radiology/Scan share; Dentist and Orthoptist carry the largest Clinical Photograph share. Second, it identifies the IO subset, where the answer choices themselves are visual; this subset is used for the positive-control analysis in Appendix~\ref{app:image_options}. These profession-level differences motivate the profession-stratified and image-type-stratified analyses in the main text.

\input{tables/T_profession_breakdown}

\section{Combined-mode accuracy with MHLW calibration}\label{app:combined_pass}

The main experiments report text-only and with-images results separately because the two subsets probe different capabilities and have different item distributions. For completeness, Table~\ref{tab:combined_pass} reports a combined-mode view that pools each model's correct answers across the scored text-only and with-images subsets within each profession. This table approximates the licensing-exam setting, where a candidate's performance is judged over the full examination rather than over modality-specific subsets.

We also report pass-threshold calibration using the official MHLW pass threshold for each profession-year cell. A model is counted as passing a cell if its combined accuracy for that profession and year meets or exceeds the corresponding MHLW threshold. This per-year calibration is stricter than comparing a pooled profession-level score with a median threshold, because thresholds vary by year in several professions. Per-year thresholds are included in the released evaluation metadata.

\input{tables/T_combined_appendix}

\begin{figure}[t]
  \centering
  \includegraphics[width=\linewidth]{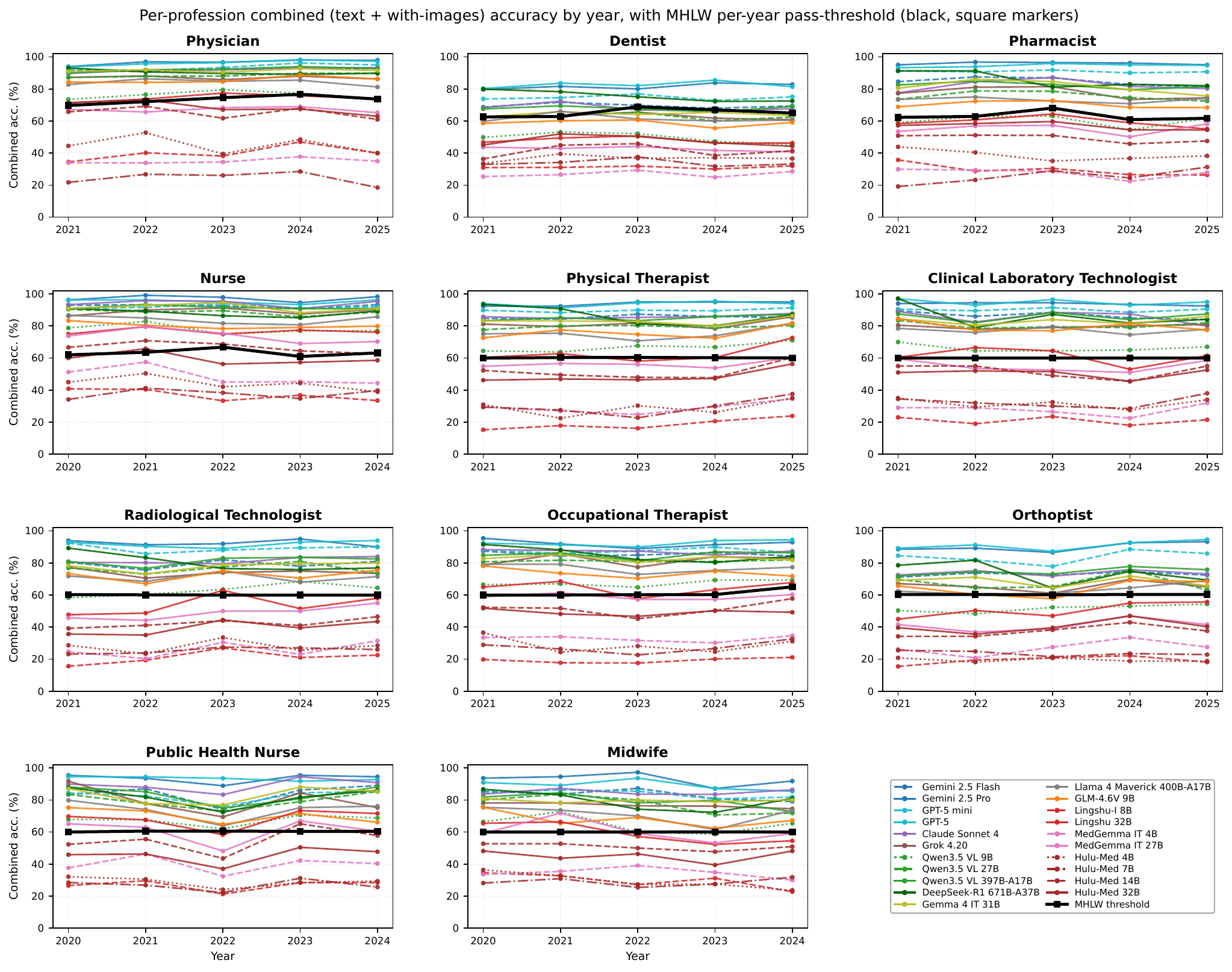}
  \caption{\textbf{Per-profession combined accuracy across the 5 covered exam years,
  with MHLW per-year pass-thresholds overlaid.} Each panel reports one profession;
  the y-axis is the combined (text-only + with-images) accuracy for that
  (model, profession, year), pooled across both modes. Year coverage differs by
  profession (Public Health Nurse, Midwife, and Nurse cover 2020--2024; the other
  eight cover 2021--2025), reflecting the JMed48k-Eval source release dates.
  Models are colour-coded by family and within-family size is encoded by line
  style (smallest = dotted; largest = solid); the bottom-right panel reproduces
  the full legend. The thick black line with square markers is the official MHLW
  pass-threshold (\texttt{threshold\_main\_pct}) for that profession-year:
  threshold variation is large in some professions (Physician 2021--2024 rises
  +7.0\,pp) and essentially flat in others (Clinical Laboratory Technologist
  fixed at 60.0\,\% in every covered year). The Pass column of
  Table~\ref{tab:combined_pass} counts, per model, the number of (profession, year)
  cells that lie at or above this black line out of $11\times 5\,=\,55$.}
  \label{fig:per-profession-over-time}
\end{figure}

Figure~\ref{fig:per-profession-over-time} disaggregates the same combined accuracy by exam year, overlaying the official MHLW per-year pass thresholds. Profession-year cells that lie at or above the threshold line correspond to the per-cell pass counts in the \emph{Pass} column of Table~\ref{tab:combined_pass}.

\section{Text-only vs.\ with-images accuracy-gap matrix}\label{app:subset_gap_matrix}

\begin{figure}[t]
  \centering
  \includegraphics[width=0.96\linewidth]{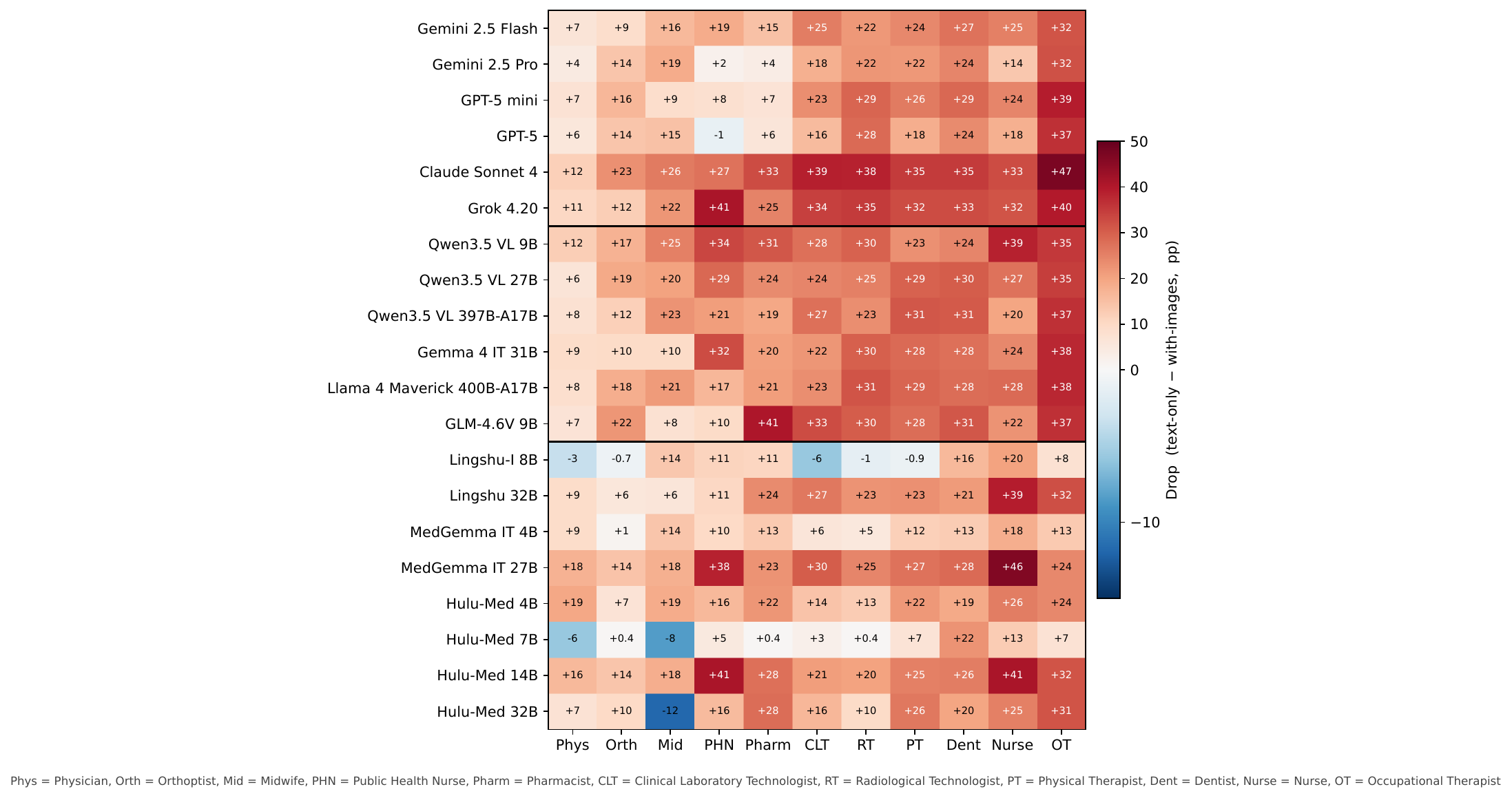}
  \caption{\textbf{Per-(model, profession) text-only versus with-images accuracy gap on JMed48k-Eval.}
  Each cell reports text-only accuracy minus with-images accuracy in percentage points for the corresponding model and profession.
  Because the two subsets contain different questions, this matrix should be interpreted as a subset-difficulty gap rather than a causal image-removal effect.
  Rows are the 20 multimodal-capable models grouped by family (Proprietary, Open-source, Medical-specific); columns are the 11 professions ordered by mean gap ascending.
  Red indicates larger text-only advantage, and blue indicates cases where with-images accuracy is higher.}
  \label{fig:subset_gap_matrix}
\end{figure}

Figure~\ref{fig:subset_gap_matrix} expands the Section~\ref{sec:image-bearing} comparison between text-only and with-images accuracy to every model and profession. Each cell reports text-only accuracy minus with-images accuracy in percentage points. Because the two subsets contain different questions, this matrix measures a subset-level difficulty gap rather than a causal effect of visual input; the causal contribution of image evidence is measured separately by the paired image-removal audit in Sections~\ref{sec:image-removal}--\ref{sec:profession-image-use}.

\section{Full paired image-removal results}\label{app:paired_audit}

This appendix collects the long-form support for the paired image-removal audit reported in Sections~\ref{sec:image-removal} and~\ref{sec:profession-image-use}: the per-model answer-transition decomposition, the per-(profession, cohort) breakdown, and the per-(model, image type) heatmap, in that order.

\subsection{Per-model answer-transition details}\label{app:transition-details}

\input{tables/T4_bcad}

Table~\ref{tab:transition-details} reports the per-model answer-transition decomposition underlying Figure~\ref{fig:image-removal-transitions}. Each row gives \(a_{\mathrm{with}}\), \(a_{\mathrm{removed}}\), \(p_{11}\), \(p_{10}\), \(p_{01}\), \(p_{00}\), and \(\Delta_{\mathrm{img}}\) on the same paired question pool, pooled across 11 professions and 5 covered exam years.

\subsection{Profession-level paired results}\label{app:profession-delta-full}

\input{tables/T_profession_va}

Table~\ref{tab:profession_delta_full} reports per-(profession, cohort) cohort-mean $a_{\mathrm{with}}$, $a_{\mathrm{removed}}$, $p_{11}$, and $\Delta_{\mathrm{img}}$. It is the long-form complement of Figure~\ref{fig:profession-combined}(b) in the main text, supporting the profession-level argument in Section~\ref{sec:profession-image-use}.

\subsection{Image-type paired results}\label{app:image-type-matrix}

\input{tables/T5_image_type}

Table~\ref{tab:image_type_delta_full} reports the same paired analysis stratified by primary image type rather than profession. Reporting $a_{\mathrm{with}}$ and $a_{\mathrm{removed}}$ separately distinguishes two qualitatively different cases that $\Delta_{\mathrm{img}}$ alone cannot separate: image types where the cohort can answer most items correctly without the image (Radiology/Scan: proprietary $a_{\mathrm{with}} = 77.3$ vs $a_{\mathrm{removed}} = 68.1$, with $\Delta_{\mathrm{img}} = +9.2$; Clinical Photograph similarly), and image types where the cohort cannot answer items at all (Medical-specific on Chemical/Molecular: $a_{\mathrm{with}} = 25.8$, $a_{\mathrm{removed}} = 18.8$, $\Delta_{\mathrm{img}} = +6.9$, with both absolute values close to the floor). The profession-level pattern in Section~\ref{sec:profession-image-use} is consistent with this image-type composition: Public Health Nurse and Pharmacist are dominated by charts and chemical structures, image types on which the proprietary cohort gains substantially; Physician and Dentist questions have a higher fraction of clinical photographs and radiology, image types on which the no-image accuracy is already high.

\begin{figure}[t]
  \centering
  \includegraphics[width=0.9\linewidth]{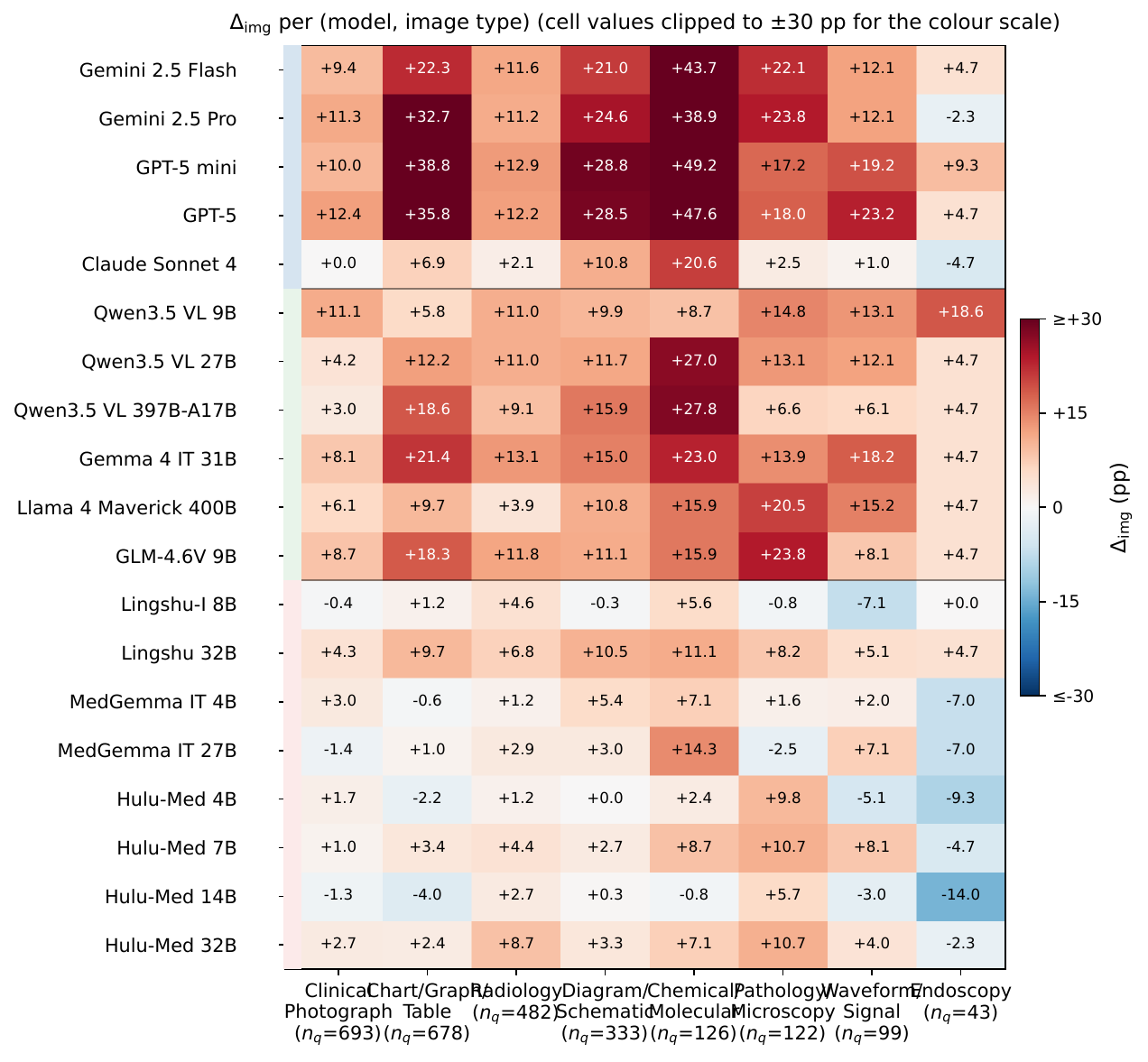}
  \caption{\textbf{Per-(model, image type) $\Delta_{\mathrm{img}}$
  heatmap.} Rows are the 20 multimodal-capable models grouped by
  family band (Proprietary, Open-source, Medical-specific). Columns
  are the eight primary image types defined in the v3.0 taxonomy
  (\S\ref{app:image_taxonomy}), ordered by sample size descending; the small
  Other/Unclear category ($n_q = 4$) is omitted. Colour encodes $\Delta_{\mathrm{img}}$ in percentage points
  on a diverging RdBu scale clipped to $\pm 30$\,pp; red cells indicate
  the image helps and blue cells indicate the image hurts. The
  cohort-mean breakdown in
  Table~\ref{tab:image_type_delta_full} collapses these rows into the
  three cohort means.}
  \label{fig:image-type-heatmap}
\end{figure}

\section{Refusal handling and the forced-guess validation}\label{app:refusal}

\input{sections/app_refusal}

\section{Image-as-Options Positive Control}\label{app:image_options}

JMed48k records, for every item with images, whether the visual content carries the answer choices themselves rather than supporting the question stem (the \texttt{image\_as\_options} flag in the released schema; Appendix~\ref{app:curation}). We refer to items with this flag set as \emph{image-as-options} (IO) items. They serve as a positive control for the visual-evidence-use audit in Section~\ref{sec:image-removal}: a model that does not read the image cannot identify any specific answer choice on these items, so accuracy is bounded above by the random baseline $1/k$ regardless of text priors. Any accuracy above this baseline must come from reading the image.

\paragraph{Why this resolves an ambiguity in \S\ref{sec:image-removal}.} The medical-specific cohort records $p_{10}$ and $p_{01}$ of comparable size in the $8$--$14\%$ range. This is consistent with two different behaviours. Images may be read but produce gains and errors that cancel, or images may not be read at all and the $16$--$20\%$ of items whose answer changes between paired runs may reflect incidental drift around the same text-prior distribution. The paired removal audit cannot distinguish these cases because both produce the same $p_{10}$, $p_{01}$, and $\Delta_{\mathrm{img}}$. The IO positive control distinguishes them.

\paragraph{Subset definition and validity checks.} The IO subset is identified by the \texttt{image\_as\_options} flag in the released metadata, which was assigned during structured extraction (Appendix~\ref{app:curation}) and verified against the official MHLW exam materials. The IO subset contains $185$ multiple-choice items in JMed48k-Eval, with $k=5$ for $170$ items and $k=4$ for $15$, giving a per-item random baseline of $\overline{1/k}=20.4\%$. A stem-leakage check (\S\ref{app:io-taxonomy}) confirms that $0$ of $185$ stems contain textual descriptions of individual option contents, so the IO assumption is not contaminated by stem-level option leakage modulo possible training-data memorization, which remains a general limitation of public examination benchmarks. An additional $38$ IO items have visually rendered numeric answers without a fixed $k$ and are reported separately in Appendix~\ref{app:io-numeric}; the cohort ordering carries over to that subset.

\begin{figure}[t]
  \centering
  \includegraphics[width=0.78\linewidth]{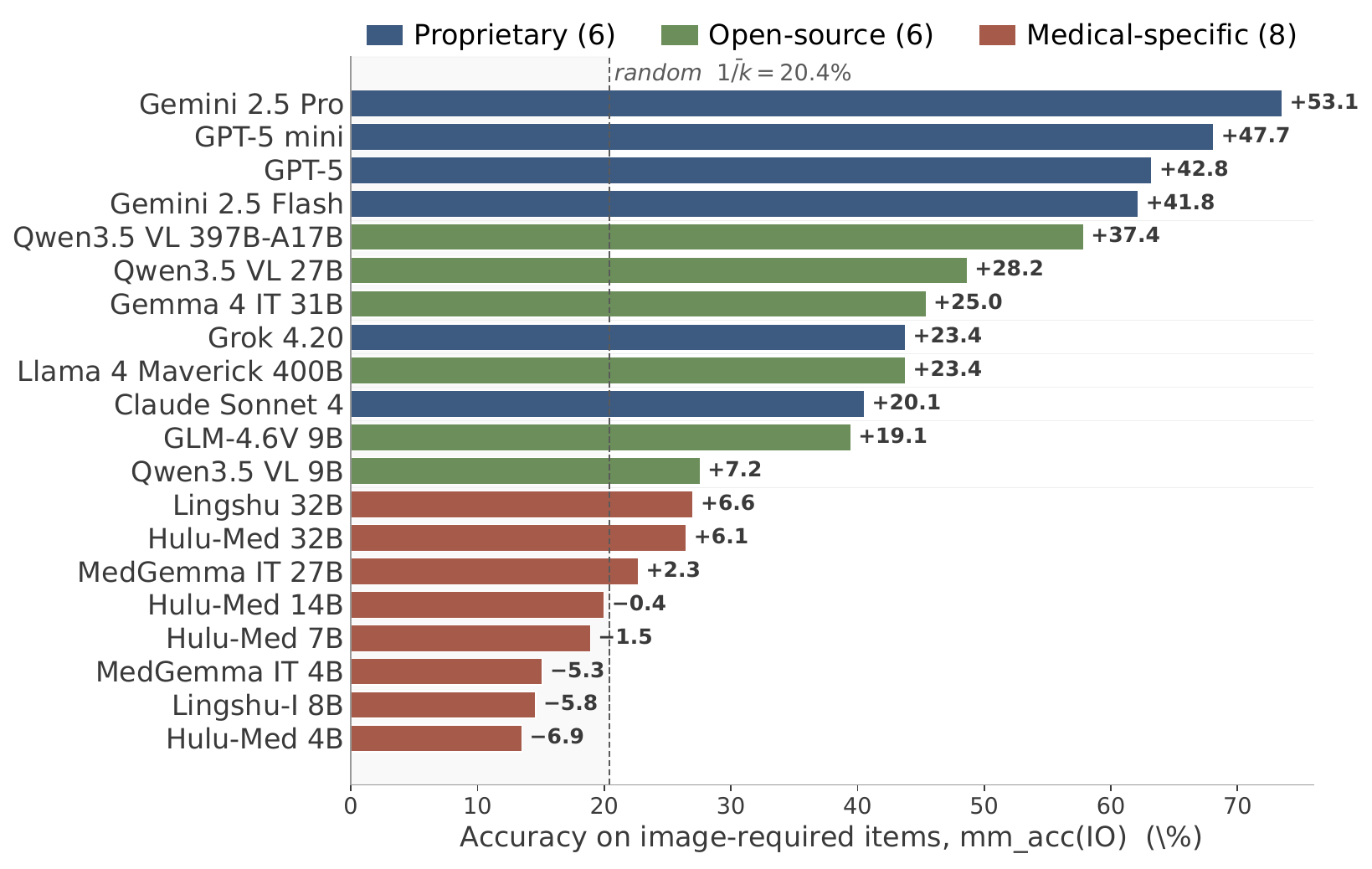}
  \caption{\textbf{Accuracy on image-required (IO) items, with
  random baseline.} Each bar is one model's $a_{\mathrm{with}}$
  on the $n_{\rm IO}=185$ MCQ items in the IO subset. The dashed
  line marks the per-item random baseline $\overline{1/k}=20.4\%$;
  the shaded zone to its left is the below-random region. Bar
  annotation gives the signed gap $a_{\mathrm{with}}-$random in
  percentage points. Any bar to the right of the baseline exceeds
  what the question stem alone can yield, since visually embedded
  answer choices remove the text-prior route to the correct option.}
  \label{fig:io}
\end{figure}

Figure~\ref{fig:io} reports per-model accuracy on the IO subset
against this baseline. The proprietary cohort averages 58.6\%, more
than 38\,pp above random. The open-source cohort averages 43.8\%,
more than 23\,pp above random. The medical-specific cohort averages
19.8\%, statistically indistinguishable from the 20.4\% random
baseline. Five of the eight medical-specific models score at or
below random.\footnote{Predicted-option distribution checks attribute
the sub-random scores of Hulu-Med~4B (13.5\%), Lingshu-I~8B (14.6\%),
MedGemma~IT~4B (15.1\%), Hulu-Med~7B (18.9\%), and Hulu-Med~14B
(20.0\%) to a mix of selection bias and output-format inconsistency,
both independent of image-derived contribution and small relative to
the $30+$ pp cohort gap. Details in Appendix~\ref{app:io-diagnostics}.}

The result resolves the ambiguity left by Section~\ref{sec:image-removal}.
The medical-specific cohort's near-random accuracy on items where
reading the image is the only path to a correct answer rules out
the ``read but cancel'' interpretation. Their $p_{10} \approx p_{01}$
in the paired removal audit reflects images not being read, with
the answer changes between paired runs consistent with stochastic
drift around the same text-prior distribution rather than
image-driven decisions in either direction.

A pharma-chemistry stress test, where the answer options are 2D
chemical structures, is reported as a subsection (\S\ref{app:io-pharma}) of this appendix.

\subsection{IO stem-leakage check}\label{app:io-taxonomy}

The IO assumption is that the question stem alone does not identify any specific answer-choice image. We checked all multiple-choice IO items for stem-level descriptions of individual options (e.g.\ patterns of the form ``$A$ is $X$, $B$ is $Y$\ldots'' or ``\textcircled{1} is $X$, \textcircled{2} is $Y$\ldots''). A regex-based search returned 0 hits. The IO assumption is therefore not contaminated by stem-level option descriptions, modulo possible training-data memorization, which remains a general limitation of public examination benchmarks.

\subsection{IO subset with numeric answers}\label{app:io-numeric}

Of the 223 IO items in the canonical scored subset, 38 have visually rendered numeric answers (e.g.\ multi-digit values rendered in the figure rather than enumerated as letter or number choices) and have no fixed answer-choice count $k$. We exclude these from the main IO analysis since the random baseline is not well defined. Cohort-mean $a_{\mathrm{with}}$ on these 38 items, by way of completeness, is $57.5\%$ (proprietary), $32.5\%$ (open-source), and $20.4\%$ (medical-specific). The cohort ordering matches that on the 185 fixed-$k$ items.

\subsection{Sub-random IO accuracy: predicted-option distribution diagnostics}\label{app:io-diagnostics}

Five medical-specific models record $a_{\mathrm{with}}$ at or below the $20.4\%$ random baseline. We checked whether these sub-random scores reflect (i) failure to extract image-derived information, (ii) selection bias in which the model over-selects a particular option label independent of the gold-option distribution, or (iii) output-format inconsistency in which the model returns labels outside the listed answer choices.

For each of the five models we computed the predicted-option distribution on the 185 MCQ IO items and compared it to the gold-option distribution. Lingshu-I~8B over-selects option A by $17$\,pp relative to the gold distribution and additionally over-selects numeric label ``1'' by another $17$\,pp; this selection bias by itself drives a substantial part of its sub-random accuracy. Hulu-Med~7B over-selects numeric label ``1'' by $12$\,pp and option B by $9$\,pp. Hulu-Med~14B over-selects option D by $9$\,pp. MedGemma~IT~4B and Hulu-Med~4B exhibit smaller letter-option deviations ($\leq 8$\,pp) but produce predictions in non-conforming formats (numeric digits, Japanese kana, or labels outside the listed choices) on a substantial fraction of items. None of these effects are correlated with image content; they are independent of $a_{\mathrm{with}}$'s image-use signal and small in magnitude relative to the $30+$\,pp cohort gap between proprietary and medical-specific cohorts.

Cohort-mean $a_{\mathrm{removed}}$ further supports the IO assumption: $19.0\%$ (proprietary), $16.7\%$ (open-source), and $15.1\%$ (medical-specific), all within $6$\,pp of the random baseline. GPT-5 mini's $a_{\mathrm{removed}}=8.1\%$ reflects calibrated refusals under image removal (Appendix~\ref{app:refusal}); Gemma 4 IT 31B's $a_{\mathrm{removed}}=6.5\%$ reflects a similar but less calibrated refusal pattern (Appendix~\ref{app:refusal}). Removing these two model-level under-shoots tightens cohort-mean $a_{\mathrm{removed}}$ to within $4$\,pp of random for all three cohorts.

\subsection{Pharma-chemistry IO stress test}\label{app:io-pharma}

\input{sections/app_io_pharma}

\section{Answer-transition case study transcripts}\label{app:cases}

\begin{CJK}{UTF8}{min}%
\input{sections/app_cases}\end{CJK}

\section{Evaluation Setup}\label{app:setup}\label{sec:setup}

This appendix documents the models, inference protocol, and accuracy metric used to produce every result in the paper.

\subsection{Models}

We benchmark 21 models grouped into three cohorts. The cohort partition is the analysis grouping used throughout the paper (Section~\ref{sec:image-removal} and Appendix~\ref{app:image-type-matrix}). DeepSeek-R1 is text-only and is included only in the text-only baseline.

\paragraph{Proprietary (6 models).} Gemini 2.5 Flash, Gemini 2.5 Pro, GPT-5 mini, GPT-5, Claude Sonnet 4, Grok 4.20.

\paragraph{Open-source general-purpose (7 models).} Qwen3.5 VL 9B, Qwen3.5 VL 27B, Qwen3.5 VL 397B-A17B, Gemma 4 IT 31B, Llama 4 Maverick 400B-A17B, GLM-4.6V 9B, and DeepSeek-R1 671B-A37B (text-only).

\paragraph{Medical-specific (8 models).} Lingshu-I 8B, Lingshu 32B, MedGemma IT 4B, MedGemma IT 27B, Hulu-Med 4B, Hulu-Med 7B, Hulu-Med 14B, Hulu-Med 32B.

For analyses that require both image-present and image-removed responses on the same items (Section~\ref{sec:image-removal}, Appendix~\ref{app:image_options}), we use the 20 multimodal-capable models, dropping DeepSeek-R1.

\begin{table}[t]
  \centering
  \caption{\textbf{Model source / provenance.} For each evaluated model
  family we cite the official technical report, system card, or model
  documentation. Citations marked ``related'' indicate the closest
  available family-level report when an exact-checkpoint report is
  unavailable.}
  \label{tab:model_provenance}
  \footnotesize
  \setlength{\tabcolsep}{6pt}
  \begin{tabular}{@{}l l l@{}}
    \toprule
    Cohort & Model family / evaluated checkpoint(s) & Reference \\
    \midrule
    Proprietary & Gemini 2.5 Flash, Gemini 2.5 Pro & \citet{comanici2025gemini25} \\
                & GPT-5, GPT-5 mini & \citet{openai2025gpt5} \\
                & Claude Sonnet 4 & \citet{anthropic2025claude4} \\
                & Grok 4.20 & \citet{xai2025grok4} \\
    \midrule
    Open-source & Qwen3.5 VL 9B / 27B / 397B-A17B & \citet{bai2025qwen3vl} (related family) \\
                & Gemma 4 IT 31B & \citet{google2026gemma4} \\
                & Llama 4 Maverick 400B-A17B & \citet{meta2025llama4} \\
                & GLM-4.6V 9B & \citet{zai2025glm46v}; \citet{hong2025glm45v} (foundation) \\
                & DeepSeek-R1 671B-A37B (text-only) & \citet{guo2025deepseekr1} \\
    \midrule
    Medical-specific & Lingshu-I 8B, Lingshu 32B & \citet{xu2025lingshu} \\
                & MedGemma IT 4B, MedGemma IT 27B & \citet{sellergren2025medgemma} \\
                & Hulu-Med 4B / 7B / 14B / 32B & \citet{jiang2025hulumed} \\
    \bottomrule
  \end{tabular}
\end{table}

\subsection{Inference protocol}

Closed-source models are queried through provider APIs at the model snapshots listed above. Open-source models are loaded under the released container image hashes and inferred with vLLM at fixed seeds. All runs use temperature $0$, single-shot prompting in Japanese, and per-model retry policies recorded in the per-cell run metadata. Reasoning is disabled where the model exposes a switch and otherwise left at the provider default.

For the paired image-removal audit, the same prompt template is used in both conditions, with the image references stripped from the message in the image-removed condition.

\subsection{Prompts}\label{app:setup_prompts}

Each question is rendered with one of two prompt templates: a text-only template for items without images, and a multimodal template for items with images. Both templates are written in Japanese and are produced by the released evaluation script in deterministic order; text below uses the released schema field names. The output schema is enforced via JSON-mode where the provider supports it, and otherwise via a JSON-format hint at the end of the prompt.

\noindent\textbf{Text-only template.}\\[2pt]
\begin{CJK}{UTF8}{min}
\noindent\fbox{\parbox{0.97\linewidth}{\footnotesize\ttfamily
\{text\_reference if present\}\\
\\
\{question\_text\}\\
\\
選択肢:\\
A. \{option a\}\\
B. \{option b\}\\
C. \{option c\}\\
D. \{option d\}\\
E. \{option e\}\\
\\
上記の質問に答えてください。\\
問題文に「2つ選べ」「3つ選べ」「すべて選べ」などの明示がない限り、解答は1つだけ選んでください。\\
複数選択が明示されている場合のみ、選んだ解答をすべて答えてください。\\
解答は、問題文中に示された選択肢のラベルのみで答えてください。\\
回答の理由も簡潔に説明してください。\\
\\
回答は以下のJSON形式で返してください：\\
\{"answer": ["解答"], "reasoning": "理由を日本語で説明"\}
}}
\end{CJK}

\noindent\textbf{Multimodal template.}\\[2pt]
The multimodal template interleaves text blocks with image blocks. The order is: optional \texttt{text\_reference}; the question stem; \texttt{content\_img} image block(s); the answer-choice list (when options are textual; for image-as-options items the choice list is omitted); \texttt{answer\_img} image block(s); the instruction block.
\begin{CJK}{UTF8}{min}
\noindent\fbox{\parbox{0.97\linewidth}{\footnotesize\ttfamily
\{text\_reference if present\}\\
\\
\{question\_text\}\\
\textit{[content image(s) inserted here]}\\
\\
選択肢:\\
A. \{option a\} \ldots E. \{option e\}\\
\textit{[answer image(s) inserted here, for image-as-options items]}\\
\\
上記の質問に答えてください。\\
問題文に「2つ選べ」「3つ選べ」「すべて選べ」などの明示がない限り、解答は1つだけ選んでください。\\
複数選択が明示されている場合のみ、選んだ解答をすべて答えてください。\\
画像・図表が含まれる場合は、それらも参照して解答してください。\\
選択肢が本文ではなく画像・図表内に示されている場合でも、解答は選択肢のラベルで答えてください。\\
解答は、問題が求める答えのみを返してください。\\
回答の理由も簡潔に説明してください。\\
\\
回答は以下のJSON形式で返してください：\\
\{"answer": ["解答"], "reasoning": "理由を日本語で説明"\}
}}
\end{CJK}

The multimodal template differs from the text-only template in three ways: the question text and option list are split by image blocks, two extra instructions remind the model to read the attached images and to answer with option labels even when the choices are themselves visual, and the final answer-format sentence is the more permissive ``\begin{CJK}{UTF8}{min}解答は、問題が求める答えのみを返してください\end{CJK}'' to accommodate items whose answer is a digit string or a layout label rather than a letter. For the paired image-removal condition (Section~\ref{sec:image-removal}), the multimodal template is used unchanged except that image blocks are dropped, leaving the same surrounding text the model would otherwise see.

\newpage
\input{checklist.tex}

\end{document}

%% file: tables/T1a_text_only.tex
% Auto-generated by paper_neurips2026/scripts/gen_t1.py -- DO NOT HAND-EDIT.
% mode=text_only
\begin{table}[t]
  \centering
  \caption{Text-only accuracy on JMed48k-Eval. \textbf{Overall} is the unweighted macro average across the 11 professions; \textcolor{red}{\textbf{best}} and \textcolor{blue}{second-best} per column. Combined-mode accuracy with MHLW pass-threshold calibration is reported in Appendix~\ref{app:combined_pass}, Table~\ref{tab:combined_pass}.}
  \label{tab:t1a}
  \footnotesize
  \setlength{\tabcolsep}{2.5pt}
  \begin{tabular}{l r @{\hskip 0.7em} r r r r r r r r r r r}
    \toprule
    \rowcolor{HeaderBand}
    Model & Overall & Phys & Dent & Pharm & Nurse & PT & CLT & RT & OT & Orth & PHN & Mid \\
    \midrule
    \rowcolor{ProprietaryBand}
    \multicolumn{13}{l}{\textbf{Proprietary}} \\
    Gemini 2.5 Flash & 86.6 & 94.0 & 81.5 & 88.3 & 93.6 & 88.7 & 91.0 & 83.2 & 88.8 & 74.2 & 85.2 & 84.4 \\
    Gemini 2.5 Pro & \textcolor{red}{\textbf{95.5}} & \textcolor{red}{\textbf{98.0}} & \textcolor{blue}{92.2} & \textcolor{red}{\textbf{96.9}} & \textcolor{red}{\textbf{97.8}} & \textcolor{red}{\textbf{96.7}} & \textcolor{blue}{96.9} & \textcolor{blue}{96.2} & \textcolor{blue}{95.2} & \textcolor{blue}{92.2} & \textcolor{red}{\textbf{93.7}} & \textcolor{red}{\textbf{94.3}} \\
    GPT-5 mini & 90.5 & 95.7 & 87.2 & 92.6 & 92.9 & 93.2 & 94.1 & 94.2 & 91.5 & 86.3 & 83.0 & 84.4 \\
    GPT-5 & \textcolor{blue}{95.1} & \textcolor{blue}{97.9} & \textcolor{red}{\textbf{92.9}} & \textcolor{blue}{96.1} & \textcolor{blue}{96.2} & \textcolor{blue}{96.2} & \textcolor{red}{\textbf{97.6}} & \textcolor{red}{\textbf{96.7}} & \textcolor{red}{\textbf{96.0}} & \textcolor{red}{\textbf{93.2}} & \textcolor{blue}{93.3} & \textcolor{blue}{90.4} \\
    Claude Sonnet 4 & 89.3 & 95.4 & 83.6 & 90.5 & 95.6 & 90.5 & 92.2 & 88.1 & 91.7 & 77.3 & 91.1 & 86.8 \\
    Grok 4.20 & 82.4 & 90.1 & 77.0 & 83.8 & 90.4 & 85.6 & 85.5 & 80.1 & 85.6 & 67.6 & 82.4 & 78.3 \\
    \addlinespace[2pt]
    \rowcolor{OpenBand}
    \multicolumn{13}{l}{\textbf{Open-source (general)}} \\
    Qwen3.5 VL 9B & 68.9 & 79.7 & 60.3 & 68.0 & 79.8 & 69.7 & 70.9 & 68.1 & 70.8 & 54.4 & 69.6 & 66.5 \\
    Qwen3.5 VL 27B & 82.0 & 90.4 & 76.0 & 81.5 & 89.8 & 83.2 & 84.7 & 80.7 & 84.6 & 70.6 & 81.7 & 78.7 \\
    Qwen3.5 VL 397B-A17B & 86.5 & 94.4 & 81.2 & 87.1 & 92.3 & 89.1 & 90.6 & 85.3 & 88.7 & 76.5 & 84.8 & 81.1 \\
    DeepSeek-R1 671B-A37B & 83.1 & 90.5 & 75.6 & 85.7 & 88.2 & 86.5 & 86.0 & 80.4 & 85.3 & 73.8 & 82.1 & 79.5 \\
    Gemma 4 IT 31B & 84.4 & 94.1 & 76.3 & 86.4 & 92.5 & 86.7 & 87.1 & 83.2 & 86.5 & 70.0 & 85.2 & 80.3 \\
    Llama 4 Maverick 400B-A17B & 77.7 & 86.5 & 73.7 & 78.4 & 85.2 & 79.2 & 81.3 & 76.4 & 80.4 & 66.3 & 75.0 & 72.4 \\
    GLM-4.6V 9B & 77.1 & 87.6 & 72.4 & 80.3 & 81.2 & 79.4 & 85.2 & 77.5 & 77.3 & 67.7 & 70.8 & 68.5 \\
    \addlinespace[2pt]
    \rowcolor{MedicalBand}
    \multicolumn{13}{l}{\textbf{Medical-specific}} \\
    Lingshu-I 8B & 27.7 & 39.0 & 38.2 & 32.3 & 37.8 & 18.6 & 20.0 & 20.8 & 20.0 & 19.1 & 27.6 & 30.7 \\
    Lingshu 32B & 65.0 & 77.0 & 57.3 & 65.5 & 78.3 & 65.8 & 65.7 & 57.7 & 67.6 & 51.5 & 68.8 & 59.6 \\
    MedGemma IT 4B & 34.0 & 37.6 & 32.5 & 30.9 & 49.4 & 30.6 & 28.8 & 26.9 & 34.1 & 27.3 & 40.4 & 35.6 \\
    MedGemma IT 27B & 60.6 & 72.1 & 55.0 & 60.9 & 75.6 & 59.8 & 60.2 & 53.2 & 61.3 & 43.4 & 63.3 & 61.8 \\
    Hulu-Med 4B & 35.8 & 50.5 & 45.0 & 44.4 & 45.2 & 31.8 & 34.1 & 30.1 & 31.3 & 20.5 & 30.0 & 30.9 \\
    Hulu-Med 7B & 29.7 & 22.7 & 43.8 & 25.6 & 38.2 & 30.4 & 33.1 & 25.6 & 28.2 & 23.7 & 27.2 & 28.1 \\
    Hulu-Med 14B & 55.2 & 69.8 & 52.6 & 56.2 & 68.4 & 54.8 & 55.5 & 45.8 & 54.5 & 39.6 & 57.6 & 52.2 \\
    Hulu-Med 32B & 53.0 & 70.6 & 56.4 & 63.8 & 60.7 & 52.0 & 53.3 & 41.3 & 52.3 & 42.0 & 46.5 & 44.3 \\
    \addlinespace[2pt]
    \bottomrule
  \end{tabular}
  \vspace{4pt}
  \parbox{\linewidth}{\footnotesize
    Profession abbreviations: Phys = Physician, Dent = Dentist, Pharm = Pharmacist, PT = Physical Therapist, CLT = Clinical Laboratory Technologist, RT = Radiological Technologist, OT = Occupational Therapist, Orth = Orthoptist, PHN = Public Health Nurse, Mid = Midwife.
  }
\end{table}

%% file: tables/T1b_with_images.tex
\begin{table}[t]
  \centering
  \caption{With-images accuracy on JMed48k-Eval ($n = 2{,}579$ scored items). Same conventions as Table~\ref{tab:t1a}; row order matches Table~\ref{tab:t1a} so models can be compared across the two subsets. DeepSeek-R1 671B-A37B is text-only and is omitted from this table.}
  \label{tab:t1b}
  \footnotesize
  \setlength{\tabcolsep}{2.5pt}
  \begin{tabular}{l r @{\hskip 0.7em} r r r r r r r r r r r}
    \toprule
    \rowcolor{HeaderBand}
    Model & Overall & Phys & Dent & Pharm & Nurse & PT & CLT & RT & OT & Orth & PHN & Mid \\
    \midrule
    \rowcolor{ProprietaryBand}
    \multicolumn{13}{l}{\textbf{Proprietary}} \\
    Gemini 2.5 Flash & 66.7 & 87.3 & 54.2 & 73.7 & 68.6 & 64.8 & 65.5 & 61.4 & 57.3 & 65.5 & 66.7 & 68.3 \\
    Gemini 2.5 Pro & \textcolor{red}{\textbf{79.7}} & \textcolor{red}{\textbf{93.7}} & \textcolor{blue}{68.0} & \textcolor{red}{\textbf{93.0}} & \textcolor{red}{\textbf{84.3}} & \textcolor{blue}{75.0} & \textcolor{blue}{78.9} & \textcolor{red}{\textbf{74.3}} & \textcolor{red}{\textbf{63.5}} & \textcolor{blue}{78.4} & \textcolor{blue}{91.7} & \textcolor{red}{\textbf{75.6}} \\
    GPT-5 mini & 70.6 & 88.6 & 58.5 & 85.8 & 68.6 & 67.2 & 70.8 & 64.9 & 52.1 & 69.8 & 75.0 & \textcolor{blue}{75.6} \\
    GPT-5 & \textcolor{blue}{78.8} & \textcolor{blue}{92.3} & \textcolor{red}{\textbf{69.3}} & \textcolor{blue}{90.0} & \textcolor{blue}{78.4} & \textcolor{red}{\textbf{78.1}} & \textcolor{red}{\textbf{81.9}} & \textcolor{blue}{68.4} & \textcolor{blue}{59.4} & \textcolor{red}{\textbf{79.3}} & \textcolor{red}{\textbf{94.4}} & 75.6 \\
    Claude Sonnet 4 & 57.7 & 83.6 & 48.8 & 57.7 & 62.7 & 55.5 & 53.2 & 49.7 & 44.8 & 54.3 & 63.9 & 61.0 \\
    Grok 4.20 & 53.6 & 79.6 & 44.5 & 58.8 & 58.8 & 53.1 & 51.5 & 45.0 & 45.8 & 55.2 & 41.7 & 56.1 \\
    \addlinespace[2pt]
    \rowcolor{OpenBand}
    \multicolumn{13}{l}{\textbf{Open-source (general)}} \\
    Qwen3.5 VL 9B & 41.8 & 67.3 & 35.8 & 37.0 & 41.2 & 46.9 & 43.3 & 38.6 & 35.4 & 37.1 & 36.1 & 41.5 \\
    Qwen3.5 VL 27B & 57.6 & 84.0 & 45.9 & 57.9 & 62.7 & 53.9 & 60.2 & 55.6 & 50.0 & 51.7 & 52.8 & 58.5 \\
    Qwen3.5 VL 397B-A17B & 63.6 & 86.8 & 50.7 & 67.9 & 72.5 & 57.8 & 63.2 & 62.0 & 52.1 & 64.7 & 63.9 & 58.5 \\
    Gemma 4 IT 31B & 61.6 & 85.0 & 48.8 & 66.0 & 68.6 & 58.6 & 64.9 & 53.2 & 49.0 & 60.3 & 52.8 & 70.7 \\
    Llama 4 Maverick 400B-A17B & 53.8 & 78.2 & 45.9 & 57.7 & 56.9 & 50.0 & 57.9 & 45.0 & 42.7 & 48.3 & 58.3 & 51.2 \\
    GLM-4.6V 9B & 52.8 & 80.6 & 41.4 & 39.8 & 58.8 & 51.6 & 52.6 & 47.4 & 40.6 & 45.7 & 61.1 & 61.0 \\
    \addlinespace[2pt]
    \rowcolor{MedicalBand}
    \multicolumn{13}{l}{\textbf{Medical-specific}} \\
    Lingshu-I 8B & 21.6 & 42.4 & 22.2 & 21.4 & 17.6 & 19.5 & 25.7 & 22.2 & 12.5 & 19.8 & 16.7 & 17.1 \\
    Lingshu 32B & 45.0 & 68.0 & 35.8 & 41.6 & 39.2 & 43.0 & 39.2 & 35.1 & 35.4 & 45.7 & 58.3 & 53.7 \\
    MedGemma IT 4B & 23.6 & 28.5 & 19.8 & 17.9 & 31.4 & 18.8 & 22.8 & 21.6 & 20.8 & 25.9 & 30.6 & 22.0 \\
    MedGemma IT 27B & 34.2 & 54.6 & 26.7 & 38.4 & 29.4 & 32.8 & 29.8 & 28.7 & 37.5 & 29.3 & 25.0 & 43.9 \\
    Hulu-Med 4B & 17.6 & 31.3 & 26.1 & 22.3 & 19.6 & 10.2 & 19.9 & 17.5 & 7.3 & 13.8 & 13.9 & 12.2 \\
    Hulu-Med 7B & 25.6 & 28.3 & 21.3 & 25.1 & 25.5 & 23.4 & 30.4 & 25.1 & 20.8 & 23.3 & 22.2 & 36.6 \\
    Hulu-Med 14B & 29.6 & 53.5 & 27.0 & 28.6 & 27.5 & 29.7 & 34.5 & 25.7 & 22.9 & 25.9 & 16.7 & 34.1 \\
    Hulu-Med 32B & 36.8 & 63.2 & 36.4 & 36.0 & 35.3 & 25.8 & 36.8 & 31.6 & 20.8 & 31.9 & 30.6 & 56.1 \\
    \addlinespace[2pt]
    \bottomrule
  \end{tabular}
  \vspace{4pt}
  \parbox{\linewidth}{\footnotesize
    Profession abbreviations: Phys = Physician, Dent = Dentist, Pharm = Pharmacist, PT = Physical Therapist, CLT = Clinical Laboratory Technologist, RT = Radiological Technologist, OT = Occupational Therapist, Orth = Orthoptist, PHN = Public Health Nurse, Mid = Midwife.
  }
\end{table}

%% file: tables/T_dataset_profession_counts.tex
% Auto-generated by paper_neurips2026/scripts/gen_dataset_tables.py -- DO NOT HAND-EDIT.
\begin{table}[t]
  \centering
  \caption{\textbf{Profession-level composition of JMed48k.} \emph{Full} columns are over the entire 2005--2025 corpus drawn from MHLW question booklets and answer materials. \emph{Scored} columns are restricted to items with a usable official gold answer and are the denominators used in all reported accuracies. The gap between the full corpus and the scored subset corresponds to the five-year split used by JMed48k-Eval and the $93$ items excluded from accuracy reporting (see Appendix~\ref{app:curation}).}
  \label{tab:dataset_profession_counts}
  \footnotesize
  \setlength{\tabcolsep}{4pt}
  \begin{tabular}{l c r r r r r r}
    \toprule
    \rowcolor{HeaderBand}
    & & \multicolumn{2}{c}{Full corpus} & \multicolumn{3}{c}{Scored JMed48k-Eval} & \\
    \rowcolor{HeaderBand}
    Profession & Years & Total & w/ img & Text-only & w/ img & Total & \% img \\
    \midrule
    Physician & 2006--2025 & 9,231 & 2,551 & 1,423 & 568 & 1,991 & 28.5\% \\
    Dentist & 2006--2025 & 7,262 & 2,956 & 992 & 771 & 1,763 & 43.7\% \\
    Pharmacist & 2012--2025 & 4,830 & 1,069 & 1,291 & 430 & 1,721 & 25.0\% \\
    Nurse & 2005--2024 & 4,320 & 175 & 1,147 & 51 & 1,198 & 4.3\% \\
    Physical Therapist & 2006--2025 & 4,000 & 629 & 859 & 128 & 987 & 13.0\% \\
    Clinical Laboratory Technologist & 2006--2025 & 4,000 & 606 & 829 & 171 & 1,000 & 17.1\% \\
    Radiological Technologist & 2006--2025 & 4,000 & 558 & 825 & 171 & 996 & 17.2\% \\
    Occupational Therapist & 2006--2025 & 3,899 & 460 & 895 & 96 & 991 & 9.7\% \\
    Orthoptist & 2006--2025 & 3,000 & 347 & 629 & 116 & 745 & 15.6\% \\
    Public Health Nurse & 2005--2024 & 2,160 & 173 & 507 & 36 & 543 & 6.6\% \\
    Midwife & 2005--2024 & 2,160 & 130 & 508 & 41 & 549 & 7.5\% \\
    \midrule
    \textbf{Total} & 2005--2025 & 48,862 & 9,654 & 9,905 & 2,579 & 12,484 & 20.7\% \\
    \bottomrule
  \end{tabular}
\end{table}

%% file: tables/T_dataset_image_types.tex
% Auto-generated by paper_neurips2026/scripts/gen_dataset_tables.py -- DO NOT HAND-EDIT.
\begin{table}[t]
  \centering
  \caption{\textbf{Image-type distribution in JMed48k.} \emph{Image-level} counts are taxonomy labels assigned to each visual element of the full-corpus images after sub-image segmentation; one source PNG may yield multiple image-level labels. \emph{Source PNG} counts assign each PNG to its dominant type. \emph{Eval Q (any)} is the number of scored JMed48k-Eval with-images questions that reference at least one image of this type; \emph{Eval Q (dom.)} assigns each question to one dominant primary type. The first eight rows are the primary inspectable visual types; \emph{Other/Unclear} is an auxiliary bucket for image references whose visual content MHLW does not publicly redistribute (typically for privacy or rights reasons), retained in released metadata but excluded from type-stratified analyses.}
  \label{tab:dataset_image_types}
  \footnotesize
  \setlength{\tabcolsep}{6pt}
  \begin{tabular}{l r r r r}
    \toprule
    \rowcolor{HeaderBand}
    Image type & Image-level & Source PNG & Eval Q (any) & Eval Q (dom.) \\
    \midrule
    Clinical Photograph & 5,832 & 3,605 & 813 & 693 \\
    Radiology/Scan & 4,934 & 3,625 & 783 & 482 \\
    Pathology/Microscopy & 1,068 & 888 & 183 & 122 \\
    Endoscopy & 268 & 206 & 45 & 43 \\
    Chart/Graph/Table & 2,599 & 2,039 & 678 & 678 \\
    Diagram/Schematic & 3,511 & 1,561 & 378 & 333 \\
    Chemical/Molecular & 1,192 & 521 & 136 & 126 \\
    Waveform/Signal & 699 & 490 & 139 & 99 \\
    \midrule
    Other/Unclear & 39 & 37 & 3 & 3 \\
    \midrule
    \textbf{Total} & 20,142 & 12,972 & 3,158 & 2,579 \\
    \bottomrule
  \end{tabular}
  \vspace{4pt}
  \parbox{\linewidth}{\footnotesize Eval Q (dom.) sums to $2{,}579$, the number of scored JMed48k-Eval with-images questions used as the denominator throughout the paper. Eval Q (any) exceeds $2{,}579$ because questions referencing multiple image types are counted once per attached type.}
\end{table}

%% file: tables/T_profession_breakdown.tex
% Auto-generated by paper_neurips2026/scripts/gen_dataset_tables.py -- DO NOT HAND-EDIT.
\begin{table}[t]
  \centering
  \caption{\textbf{Per-profession breakdown of the scored JMed48k-Eval subset.} Columns CP, RAD, PAT, END, CGT, DIA, CHM, and WAV report the dominant primary image type for scored questions with images: Clinical Photograph, Radiology/Scan, Pathology/Microscopy, Endoscopy, Chart/Graph/Table, Diagram/Schematic, Chemical/Molecular, and Waveform/Signal. \emph{Total img} sums these eight columns plus the auxiliary Other/Unclear bucket. \emph{IO} is the number of image-as-options items, where the answer choices themselves are visual. \emph{Single}, \emph{Multi}, and \emph{Numeric} report the answer-format counts used by the scorer (Appendix~\ref{app:scoring}); their sum equals the per-profession scored question count, and the table totals match the $2{,}579$ scored with-images and $12{,}484$ scored question denominators used throughout the paper.}
  \label{tab:profession_breakdown}
  \footnotesize
  \setlength{\tabcolsep}{3pt}
  \begin{tabular}{l r r r r r r r r r r r r r}
    \toprule
    \rowcolor{HeaderBand}
    & \multicolumn{8}{c}{Image type (dominant)} & & & & & \\
    \rowcolor{HeaderBand}
    Profession & CP & RAD & PAT & END & CGT & DIA & CHM & WAV & Total img & IO & Single & Multi & Numeric \\
    \midrule
    Phys & 99 & 258 & 36 & 39 & 95 & 11 & 0 & 28 & 568 & 20 & 1,725 & 250 & 16 \\
    Dent & 497 & 50 & 21 & 3 & 156 & 28 & 2 & 13 & 771 & 37 & 1,022 & 727 & 14 \\
    Pharm & 2 & 0 & 1 & 0 & 225 & 72 & 124 & 6 & 430 & 83 & 913 & 808 & 0 \\
    Nurse & 7 & 3 & 0 & 0 & 16 & 22 & 0 & 3 & 51 & 20 & 1,107 & 84 & 7 \\
    PT & 5 & 29 & 0 & 0 & 17 & 71 & 0 & 6 & 128 & 11 & 855 & 131 & 1 \\
    CLT & 16 & 20 & 63 & 1 & 35 & 10 & 0 & 26 & 171 & 6 & 825 & 175 & 0 \\
    RT & 9 & 96 & 1 & 0 & 35 & 29 & 0 & 1 & 171 & 9 & 883 & 112 & 1 \\
    OT & 13 & 13 & 0 & 0 & 9 & 58 & 0 & 3 & 96 & 21 & 857 & 131 & 3 \\
    Orth & 44 & 8 & 0 & 0 & 35 & 22 & 0 & 7 & 116 & 13 & 630 & 114 & 1 \\
    PHN & 0 & 0 & 0 & 0 & 35 & 1 & 0 & 0 & 36 & 1 & 451 & 84 & 8 \\
    Mid & 1 & 5 & 0 & 0 & 20 & 9 & 0 & 6 & 41 & 2 & 451 & 98 & 0 \\
    \midrule
    \textbf{Total} & 693 & 482 & 122 & 43 & 678 & 333 & 126 & 99 & 2,579 & 223 & 9,719 & 2,714 & 51 \\
    \bottomrule
  \end{tabular}
  \vspace{4pt}
  \parbox{\linewidth}{\footnotesize
    Profession abbreviations: Phys = Physician, Dent = Dentist, Pharm = Pharmacist, Nurse = Nurse, PT = Physical Therapist, CLT = Clinical Laboratory Technologist, RT = Radiological Technologist, OT = Occupational Therapist, Orth = Orthoptist, PHN = Public Health Nurse, Mid = Midwife.
  }
\end{table}

%% file: tables/T_combined_appendix.tex
% Auto-generated by paper_neurips2026/scripts/gen_t_combined.py
\begin{table}[t]
  \centering
  \caption{\textbf{Combined text+with-images accuracy} on the scored JMed48k-Eval subset. Each profession cell pools the model's correct answers across the text-only and with-images subsets over the five covered exam years. \textbf{Overall} is the unweighted macro average across the 11 professions. \textbf{Pass} counts the number of profession-year cells, out of 55, in which the model's combined accuracy meets or exceeds that year's official MHLW pass threshold. The bottom row reports each profession's median MHLW threshold across the five years; the \textbf{Pass} column shows the 55-cell denominator. Per-year thresholds are included in the released evaluation metadata. \textcolor{red}{\textbf{Best}} and \textcolor{blue}{second-best} per column.
  }
  \label{tab:combined_pass}
  \footnotesize
  \setlength{\tabcolsep}{2.5pt}
  \begin{tabular}{l r r @{\hskip 0.7em} r r r r r r r r r r r}
    \toprule
    \rowcolor{HeaderBand}
    Model & Overall & Pass & Phys & Dent & Pharm & Nurse & PT & CLT & RT & OT & Orth & PHN & Mid \\
    \midrule
    \rowcolor{ProprietaryBand}
    \multicolumn{14}{l}{\textbf{Proprietary}} \\
    Gemini 2.5 Flash & 83.3 & \textcolor{red}{\textbf{55}} & 92.1 & 69.5 & 84.7 & 92.6 & 85.6 & 86.6 & 79.4 & 85.8 & 72.9 & 84.0 & 83.2 \\
    Gemini 2.5 Pro & \textcolor{red}{\textbf{92.8}} & \textcolor{red}{\textbf{55}} & \textcolor{red}{\textbf{96.7}} & \textcolor{blue}{81.6} & \textcolor{red}{\textbf{95.9}} & \textcolor{red}{\textbf{97.2}} & \textcolor{red}{\textbf{93.9}} & \textcolor{blue}{93.8} & \textcolor{red}{\textbf{92.5}} & \textcolor{blue}{92.1} & \textcolor{blue}{90.1} & \textcolor{red}{\textbf{93.6}} & \textcolor{red}{\textbf{92.9}} \\
    GPT-5 mini & 87.1 & \textcolor{red}{\textbf{55}} & 93.7 & 74.6 & 90.9 & 91.9 & 89.9 & 90.1 & 89.2 & 87.7 & 83.8 & 82.5 & 83.8 \\
    GPT-5 & \textcolor{blue}{92.3} & \textcolor{red}{\textbf{55}} & \textcolor{blue}{96.3} & \textcolor{red}{\textbf{82.6}} & \textcolor{blue}{94.6} & \textcolor{blue}{95.4} & \textcolor{blue}{93.8} & \textcolor{red}{\textbf{94.9}} & \textcolor{blue}{91.9} & \textcolor{red}{\textbf{92.4}} & \textcolor{red}{\textbf{91.0}} & \textcolor{blue}{93.4} & \textcolor{blue}{89.3} \\
    Claude Sonnet 4 & 84.1 & 53 & 92.0 & 68.3 & 82.3 & 94.2 & 85.9 & 85.5 & 81.5 & 87.2 & 73.7 & 89.3 & 84.9 \\
    Grok 4.20 & 77.8 & 51 & 87.1 & 62.8 & 77.6 & 89.1 & 81.4 & 79.7 & 74.1 & 81.7 & 65.6 & 79.7 & 76.7 \\
    \addlinespace[2pt]
    \rowcolor{OpenBand}
    \multicolumn{14}{l}{\textbf{Open-source (general)}} \\
    Qwen3.5 VL 9B & 64.7 & 39 & 76.1 & 49.6 & 60.3 & 78.1 & 66.8 & 66.2 & 63.1 & 67.4 & 51.7 & 67.4 & 64.7 \\
    Qwen3.5 VL 27B & 78.0 & 52 & 88.5 & 62.8 & 75.6 & 88.6 & 79.4 & 80.5 & 76.4 & 81.2 & 67.7 & 79.7 & 77.2 \\
    Qwen3.5 VL 397B-A17B & 82.6 & 53 & 92.2 & 67.9 & 82.3 & 91.5 & 85.0 & 85.9 & 81.3 & 85.2 & 74.6 & 83.4 & 79.4 \\
    DeepSeek-R1 671B-A37B & 83.1 & \textcolor{red}{\textbf{55}} & 90.5 & 75.6 & 85.7 & 88.2 & 86.5 & 86.0 & 80.4 & 85.3 & 73.8 & 82.1 & 79.5 \\
    Gemma 4 IT 31B & 80.6 & 51 & 91.5 & 64.3 & 81.3 & 91.5 & 83.1 & 83.3 & 78.0 & 82.8 & 68.5 & 83.1 & 79.6 \\
    Llama 4 Maverick 400B-A17B & 73.8 & 51 & 84.1 & 61.5 & 73.2 & 84.0 & 75.4 & 77.3 & 71.0 & 76.8 & 63.5 & 73.8 & 70.9 \\
    GLM-4.6V 9B & 72.6 & 49 & 85.6 & 58.8 & 70.2 & 80.2 & 75.8 & 79.6 & 72.3 & 73.8 & 64.3 & 70.2 & 67.9 \\
    \addlinespace[2pt]
    \rowcolor{MedicalBand}
    \multicolumn{14}{l}{\textbf{Medical-specific}} \\
    Lingshu-I 8B & 26.7 & 0 & 40.0 & 31.2 & 29.6 & 37.0 & 18.7 & 21.0 & 21.1 & 19.3 & 19.2 & 26.9 & 29.7 \\
    Lingshu 32B & 61.7 & 28 & 74.4 & 47.9 & 59.5 & 76.6 & 62.8 & 61.2 & 53.8 & 64.5 & 50.6 & 68.1 & 59.2 \\
    MedGemma IT 4B & 32.3 & 0 & 35.0 & 26.9 & 27.7 & 48.7 & 29.1 & 27.8 & 26.0 & 32.8 & 27.1 & 39.8 & 34.6 \\
    MedGemma IT 27B & 56.4 & 12 & 67.1 & 42.7 & 55.3 & 73.6 & 56.3 & 55.0 & 49.0 & 59.0 & 41.2 & 60.8 & 60.5 \\
    Hulu-Med 4B & 32.7 & 0 & 45.1 & 36.7 & 38.9 & 44.2 & 29.0 & 31.7 & 27.9 & 29.0 & 19.5 & 28.9 & 29.5 \\
    Hulu-Med 7B & 28.7 & 0 & 24.3 & 33.9 & 25.5 & 37.6 & 29.5 & 32.6 & 25.5 & 27.4 & 23.6 & 26.9 & 28.8 \\
    Hulu-Med 14B & 51.2 & 6 & 65.1 & 41.4 & 49.3 & 66.7 & 51.6 & 51.9 & 42.4 & 51.5 & 37.4 & 54.9 & 50.8 \\
    Hulu-Med 32B & 50.2 & 3 & 68.5 & 47.6 & 56.9 & 59.6 & 48.6 & 50.5 & 39.7 & 49.2 & 40.4 & 45.5 & 45.2 \\
    \addlinespace[2pt]
    \midrule
    \rowcolor{MHLWBand}
    MHLW threshold & \multicolumn{1}{c}{--} & 55 & 73.7 & 65.0 & 62.3 & 63.2 & 60.2 & 60.0 & 60.0 & 60.2 & 60.4 & 60.4 & 60.0 \\
    \bottomrule
  \end{tabular}
  \vspace{4pt}
  \parbox{\linewidth}{\footnotesize
    Profession abbreviations: Phys = Physician, Dent = Dentist, Pharm = Pharmacist, Nurse = Nurse, PT = Physical Therapist, CLT = Clinical Laboratory Technologist, RT = Radiological Technologist, OT = Occupational Therapist, Orth = Orthoptist, PHN = Public Health Nurse, Mid = Midwife.\\
    \textbf{MHLW threshold computation.} For each profession we take the median across the 5 covered exam years of \texttt{threshold\_main\_pct} from our MHLW calibration file (55 rows, one per profession-year). \texttt{threshold\_main\_pct} is the percentage form of the official MHLW academic-difficulty pass cut-off (the \textit{ippan + rinsho} component for the Physician exam, the \textit{soutoutoten} component for allied health). It excludes the separate \textit{hisshu} (must-not-fail) $\geq$80\% sub-gate, which we cannot reconstruct because our items are not labelled with the hisshu/ippan/rinsho partition.
  }
\end{table}

%% file: tables/T4_bcad.tex
% Auto-generated by paper_neurips2026/scripts/gen_t4_bcad.py
\begin{table}[t]
  \centering
  \caption{\textbf{Per-model answer-transition decomposition on the with-images scored subset of JMed48k-Eval.} For each item, the model's response with the image is paired with its response after image removal. Items partition into four answer-transition fractions, $p_{11}$ correct in both, $p_{10}$ correct only with the image, $p_{01}$ correct only without the image, $p_{00}$ wrong in both. The net image effect is $\Delta_{\mathrm{img}} = p_{10} - p_{01}$, equivalent to $a_{\mathrm{with}} - a_{\mathrm{removed}}$. $p_{ij}$ values are percentages of the per-model paired item pool, $a_{\mathrm{with}}$ and $a_{\mathrm{removed}}$ are accuracies with and without the image. The largest positive $\Delta_{\mathrm{img}}$ is shown in \textcolor{red}{\textbf{red}}, the second largest in \textcolor{blue}{blue}, negative values in \textcolor{red}{red}.}
  \label{tab:transition-details}
  \footnotesize
  \setlength{\tabcolsep}{4pt}
  \begin{tabular}{l r r r r r r r r}
    \toprule
    \rowcolor{HeaderBand}
    Model & $n$ & $a_{\mathrm{with}}$ & $a_{\mathrm{removed}}$ & $p_{11}\%$ & $p_{10}\%$ & $p_{01}\%$ & $p_{00}\%$ & $\Delta_{\mathrm{img}}$ \\
    \midrule
    \rowcolor{ProprietaryBand}
    \multicolumn{9}{l}{\textbf{Proprietary}} \\
    Gemini 2.5 Flash & 2579 & 67.8 & 50.8 & 45.6 & 22.2 & 5.2 & 26.9 & +17.0 \\
    Gemini 2.5 Pro & 2579 & 80.4 & 60.0 & 55.5 & 24.9 & 4.5 & 15.1 & +20.4 \\
    GPT-5 mini & 2579 & 72.3 & 49.2 & 44.6 & 27.7 & 4.6 & 23.1 & \textcolor{red}{\textbf{+23.1}} \\
    GPT-5 & 2579 & 79.7 & 56.8 & 53.2 & 26.6 & 3.7 & 16.6 & \textcolor{blue}{+22.9} \\
    Claude Sonnet 4 & 2579 & 59.4 & 54.7 & 47.2 & 12.2 & 7.5 & 33.1 & +4.7 \\
    Grok 4.20 & 2579 & 56.5 & 49.6 & 44.1 & 12.4 & 5.5 & 38.0 & +6.9 \\
    \addlinespace[2pt]
    \rowcolor{OpenBand}
    \multicolumn{9}{l}{\textbf{Open-source (general)}} \\
    Qwen3.5 VL 9B & 2579 & 44.4 & 34.6 & 27.0 & 17.4 & 7.6 & 48.0 & +9.8 \\
    Qwen3.5 VL 27B & 2579 & 59.3 & 48.9 & 43.1 & 16.2 & 5.9 & 34.8 & +10.4 \\
    Qwen3.5 VL 397B-A17B & 2579 & 64.9 & 53.4 & 43.9 & 21.0 & 9.5 & 25.6 & +11.4 \\
    Gemma 4 IT 31B & 2579 & 62.8 & 48.1 & 42.4 & 20.4 & 5.7 & 31.5 & +14.7 \\
    Llama 4 Maverick 400B-A17B & 2579 & 56.4 & 47.7 & 39.4 & 17.0 & 8.3 & 35.4 & +8.7 \\
    GLM-4.6V 9B & 2579 & 52.5 & 39.4 & 31.8 & 20.7 & 7.7 & 39.8 & +13.1 \\
    \addlinespace[2pt]
    \rowcolor{MedicalBand}
    \multicolumn{9}{l}{\textbf{Medical-specific}} \\
    Lingshu-I 8B & 2579 & 25.9 & 25.0 & 15.9 & 10.0 & 9.2 & 64.9 & +0.9 \\
    Lingshu 32B & 2579 & 45.5 & 38.0 & 31.7 & 13.8 & 6.3 & 48.2 & +7.5 \\
    MedGemma IT 4B & 2579 & 22.4 & 20.4 & 12.0 & 10.4 & 8.4 & 69.2 & +2.0 \\
    MedGemma IT 27B & 2579 & 36.3 & 34.8 & 25.4 & 10.9 & 9.4 & 54.4 & +1.5 \\
    Hulu-Med 4B & 2579 & 23.1 & 22.8 & 14.8 & 8.3 & 8.0 & 68.9 & +0.3 \\
    Hulu-Med 7B & 2579 & 24.9 & 21.4 & 12.7 & 12.2 & 8.7 & 66.5 & +3.5 \\
    Hulu-Med 14B & 2579 & 33.4 & 34.4 & 24.7 & 8.7 & 9.7 & 56.8 & \textcolor{red}{-1.0} \\
    Hulu-Med 32B & 2579 & 40.9 & 36.5 & 29.3 & 11.6 & 7.3 & 51.9 & +4.3 \\
    \addlinespace[2pt]
    \bottomrule
  \end{tabular}
\end{table}

%% file: tables/T_profession_va.tex
% Auto-generated by paper_neurips2026/scripts/gen_t_profession_va.py
\begin{table}[t]
  \centering
  \caption{\textbf{Profession-level paired image-removal results.} For each profession and cohort, $a_{\mathrm{with}}$ is the cohort-mean accuracy in the original setting, $a_{\mathrm{removed}}$ is the cohort-mean accuracy after image removal on the same questions, and $p_{11}$ is the share of questions answered correctly in both settings. All values except $n$ are reported in percentages, with $\Delta_{\mathrm{img}} = a_{\mathrm{with}} - a_{\mathrm{removed}}$ in percentage points. Figure~\ref{fig:profession-combined}(b) visualises the $\Delta_{\mathrm{img}}$ column.}
  \label{tab:profession_delta_full}
  \footnotesize
  \setlength{\tabcolsep}{5pt}
  \begin{tabular}{l r l r r r r}
    \toprule
    \rowcolor{HeaderBand}
    Profession & $n$ & Cohort & $a_{\mathrm{with}}$ & $a_{\mathrm{removed}}$ & $p_{11}$ & $\Delta_{\mathrm{img}}$ \\
    \midrule
    Physician & 568 & Proprietary & 87.5 & 81.8 & 78.7 & +5.7 \\
     &  & Open-source & 80.3 & 71.5 & 66.8 & +8.8 \\
     &  & Medical-specific & 46.2 & 47.0 & 36.3 & $-$0.8 \\
    \addlinespace[1pt]
    Dentist & 771 & Proprietary & 57.2 & 49.2 & 42.5 & +8.0 \\
     &  & Open-source & 44.7 & 37.8 & 29.8 & +6.9 \\
     &  & Medical-specific & 26.9 & 26.3 & 18.0 & +0.6 \\
    \addlinespace[1pt]
    Orthoptist & 116 & Proprietary & 67.1 & 54.5 & 48.7 & +12.6 \\
     &  & Open-source & 51.3 & 46.6 & 36.4 & +4.7 \\
     &  & Medical-specific & 26.9 & 23.8 & 17.1 & +3.1 \\
    \addlinespace[1pt]
    Midwife & 41 & Proprietary & 68.7 & 54.9 & 50.0 & +13.8 \\
     &  & Open-source & 56.9 & 47.6 & 37.4 & +9.3 \\
     &  & Medical-specific & 34.5 & 31.7 & 21.6 & +2.7 \\
    \addlinespace[1pt]
    Occupational Therapist & 96 & Proprietary & 53.8 & 38.9 & 31.4 & +14.9 \\
     &  & Open-source & 45.0 & 37.2 & 27.1 & +7.8 \\
     &  & Medical-specific & 22.3 & 18.8 & 12.1 & +3.5 \\
    \addlinespace[1pt]
    Physical Therapist & 128 & Proprietary & 65.6 & 46.7 & 41.8 & +18.9 \\
     &  & Open-source & 53.1 & 41.1 & 32.3 & +12.0 \\
     &  & Medical-specific & 25.4 & 20.6 & 14.4 & +4.8 \\
    \addlinespace[1pt]
    Clinical Laboratory Technologist & 171 & Proprietary & 67.0 & 43.4 & 36.5 & +23.6 \\
     &  & Open-source & 57.0 & 38.0 & 29.5 & +19.0 \\
     &  & Medical-specific & 29.9 & 22.4 & 14.6 & +7.5 \\
    \addlinespace[1pt]
    Radiological Technologist & 171 & Proprietary & 60.6 & 33.3 & 27.2 & +27.3 \\
     &  & Open-source & 50.3 & 33.8 & 24.4 & +16.5 \\
     &  & Medical-specific & 26.0 & 20.0 & 12.5 & +5.9 \\
    \addlinespace[1pt]
    Pharmacist & 430 & Proprietary & 76.5 & 45.0 & 41.2 & +31.6 \\
     &  & Open-source & 54.4 & 37.4 & 30.3 & +17.0 \\
     &  & Medical-specific & 28.9 & 24.7 & 17.3 & +4.2 \\
    \addlinespace[1pt]
    Nurse & 51 & Proprietary & 70.3 & 35.0 & 29.1 & +35.3 \\
     &  & Open-source & 60.1 & 29.1 & 22.5 & +31.0 \\
     &  & Medical-specific & 28.2 & 20.3 & 11.3 & +7.8 \\
    \addlinespace[1pt]
    Public Health Nurse & 36 & Proprietary & 72.2 & 32.4 & 30.1 & +39.8 \\
     &  & Open-source & 54.2 & 32.9 & 28.2 & +21.3 \\
     &  & Medical-specific & 26.7 & 21.9 & 16.3 & +4.9 \\
    \addlinespace[1pt]
    \bottomrule
  \end{tabular}
\end{table}

%% file: tables/T5_image_type.tex
% Auto-generated by paper_neurips2026/scripts/gen_t5_image_type.py
\begin{table}[t]
  \centering
  \caption{\textbf{Image-type paired image-removal results.} For each primary image type and cohort, we report $a_{\mathrm{with}}$, $a_{\mathrm{removed}}$, $p_{11}$, and $\Delta_{\mathrm{img}}$ on the same set of questions. All values except $n_q$ are reported in percentages, with $\Delta_{\mathrm{img}}$ in percentage points. Image types are ordered by the number of questions. The small Other/Unclear category ($n_q = 4$) is omitted.}
  \label{tab:image_type_delta_full}
  \footnotesize
  \setlength{\tabcolsep}{5pt}
  \begin{tabular}{l r l r r r r}
    \toprule
    \rowcolor{HeaderBand}
    Image type & $n_q$ & Cohort & $a_{\mathrm{with}}$ & $a_{\mathrm{removed}}$ & $p_{11}$ & $\Delta_{\mathrm{img}}$ \\
    \midrule
    Clinical Photograph & 693 & Proprietary & 62.7 & 55.1 & 49.0 & +7.6 \\
     &  & Open-source & 51.2 & 44.3 & 36.7 & +6.9 \\
     &  & Medical-specific & 30.0 & 28.8 & 20.7 & +1.2 \\
    \addlinespace[1pt]
    Chart/Graph/Table & 678 & Proprietary & 69.5 & 45.1 & 40.0 & +24.4 \\
     &  & Open-source & 51.9 & 37.6 & 29.1 & +14.3 \\
     &  & Medical-specific & 26.5 & 25.1 & 16.6 & +1.4 \\
    \addlinespace[1pt]
    Radiology/Scan & 482 & Proprietary & 77.3 & 68.1 & 63.8 & +9.2 \\
     &  & Open-source & 71.0 & 61.0 & 55.6 & +10.0 \\
     &  & Medical-specific & 41.9 & 37.8 & 29.7 & +4.1 \\
    \addlinespace[1pt]
    Diagram/Schematic & 333 & Proprietary & 62.1 & 42.1 & 36.3 & +20.0 \\
     &  & Open-source & 48.7 & 36.3 & 26.7 & +12.4 \\
     &  & Medical-specific & 24.8 & 21.7 & 13.9 & +3.1 \\
    \addlinespace[1pt]
    Chemical/Molecular & 126 & Proprietary & 75.1 & 37.0 & 35.1 & +38.1 \\
     &  & Open-source & 49.7 & 30.0 & 24.6 & +19.7 \\
     &  & Medical-specific & 25.8 & 18.8 & 12.4 & +6.9 \\
    \addlinespace[1pt]
    Pathology/Microscopy & 122 & Proprietary & 80.1 & 64.9 & 59.8 & +15.2 \\
     &  & Open-source & 70.5 & 55.1 & 49.0 & +15.4 \\
     &  & Medical-specific & 40.4 & 34.9 & 24.9 & +5.4 \\
    \addlinespace[1pt]
    Waveform/Signal & 99 & Proprietary & 74.6 & 62.3 & 56.4 & +12.3 \\
     &  & Open-source & 67.7 & 55.6 & 47.5 & +12.1 \\
     &  & Medical-specific & 38.5 & 37.1 & 27.1 & +1.4 \\
    \addlinespace[1pt]
    Endoscopy & 43 & Proprietary & 80.2 & 79.5 & 73.6 & +0.8 \\
     &  & Open-source & 78.3 & 71.3 & 66.3 & +7.0 \\
     &  & Medical-specific & 48.0 & 52.9 & 39.2 & $-$4.9 \\
    \addlinespace[1pt]
    \bottomrule
  \end{tabular}
\end{table}

%% file: sections/app_refusal.tex
% app:refusal --- Refusal-handling alternatives and forced-guess validation

\subsection*{Refusal-handling alternatives}

A small fraction of GPT-5 family responses under vision ablation are
calibrated refusals rather than parse failures.  The model returns a
well-formed JSON object but leaves the \texttt{answer} array empty,
with a Japanese-language explanation such as
\textit{``\begin{CJK}{UTF8}{min}別冊No.Xが提示されていないため回答できません\end{CJK}''}
(``I cannot answer because the attached figure is not provided'').
These refusals are qualitatively distinct from API errors: the error
field is empty, and the reasoning field contains a coherent
explanation of why the model declines.

We report $\Delta_{\mathrm{img}}$ strict, counting all refusals as
incorrect; this is the most conservative and reviewer-interpretable
choice.  Two alternatives appear in
Table~\ref{tab:dimg-alternatives}:
\emph{conditional $\Delta_{\mathrm{img}}$} excludes refused items
from both numerator and denominator; \emph{imputed
$\Delta_{\mathrm{img}}$} fills refusals with the per-model
forced-guess accuracy measured in the validation probe below.

\begin{table}[h]
\centering\footnotesize
\caption{$\Delta_{\mathrm{img}}$ under three refusal-handling
conventions for the GPT-5 family (refusal rates are essentially zero
for all other models in our suite). Strict = refusal counted as
wrong; Conditional = refused items excluded from denominator;
Imputed = refused items filled with the per-model forced-guess
accuracy.}
\label{tab:dimg-alternatives}
\begin{tabular}{l r r r r}
\toprule
Model & Refusal rate & Strict & Conditional & Imputed\\
\midrule
GPT-5           & 11.2\% & $+22.9$\,pp & $+17.0$\,pp & $+20.1$\,pp\\
GPT-5 mini      & 20.3\% & $+23.1$\,pp & $+12.2$\,pp & $+18.1$\,pp\\
\bottomrule
\end{tabular}
\end{table}

\noindent
Strict and conditional $\Delta_{\mathrm{img}}$ differ substantially
for GPT-5 mini ($\sim$11\,pp) and moderately for GPT-5 ($\sim$6\,pp).  Because
the forced-guess validation below confirms that refusals reflect
genuine epistemic uncertainty, the conditional and imputed
conventions better represent the model's actual use of visual
evidence; we report both in this appendix to bound the true
$\Delta_{\mathrm{img}}$.  The strict convention (main paper) is the
most conservative choice and is consistent with how all other models
are evaluated.

\subsection*{Forced-guess validation}

To confirm that GPT-5 family refusals reflect genuine uncertainty rather
than safety over-conservatism, we re-issued the 746 refused vision-ablation
items with an explicit instruction permitting guessing:
``\textit{Even if you cannot see the figure, please choose one answer based
on your best guess.}''  Table~\ref{tab:forced-guess} reports accuracy
on the items that received a prediction after the forced-guess prompt.

\begin{table}[h]
\centering\footnotesize
\caption{Forced-guess results.  Items that refused under standard
vision-ablation are re-issued with an explicit permission-to-guess
instruction.  Accuracy near the random baseline ($\approx$20\%) confirms
the refusals reflect genuine uncertainty, not safety over-conservatism.
The refusal counts in this table were measured at the time the
forced-guess probe was run; the canonical paired set used in
Table~\ref{tab:dimg-alternatives} (after a subsequent
parser-coverage update) contains $289$ and $524$ refusals for
GPT-5 and GPT-5 mini, respectively.  The qualitative finding
($\approx$25\% accuracy, near the 20\% random baseline) is
unchanged.}
\label{tab:forced-guess}
\begin{tabular}{l r r r r}
\toprule
Model & Refused (total) & Parsed ($n$) & Forced-guess acc. & Random baseline\\
\midrule
GPT-5         & 260 & 250 (96.2\%) & \textbf{25.0\%} & $\approx$20\%\\
GPT-5 mini    & 486 & 475 (97.7\%) & \textbf{24.5\%} & $\approx$20\%\\
\bottomrule
\end{tabular}
\end{table}

\noindent
Both models answer at near-random accuracy when forced to guess,
confirming the refusals are epistemically calibrated: the model
correctly identifies that the image is necessary and that it cannot
answer from text alone.  This validates the ``calibrated refusal''
characterisation in the main text (\S\ref{app:image_options}).

\subsection*{Gemma 4 IT 31B: a similar but less calibrated pattern}

Gemma 4 IT 31B exhibits a refusal pattern under image removal that
resembles the GPT-5 family but is less calibrated. On the IO subset
($n = 121$), $\sim 22\%$ of its $a_{\mathrm{removed}}$ predictions are
null and a further $\sim 22\%$ contain short Japanese refusal templates
such as
\textit{``\begin{CJK}{UTF8}{min}画像がないため判定不可\end{CJK}''}
(``cannot determine without image''). Unlike GPT-5 family refusals,
Gemma's refusal text is less consistent and the refusal rate is not
estimated outside the IO subset; we therefore do not apply the
conditional or imputed $\Delta_{\mathrm{img}}$ conventions to Gemma.
Its strict $\Delta_{\mathrm{img}}$ is reported in
Table~\ref{tab:transition-details} alongside other models without
refusal correction.

%% file: sections/app_io_pharma.tex
\begin{figure}[h]
  \centering
  \includegraphics[width=0.85\linewidth]{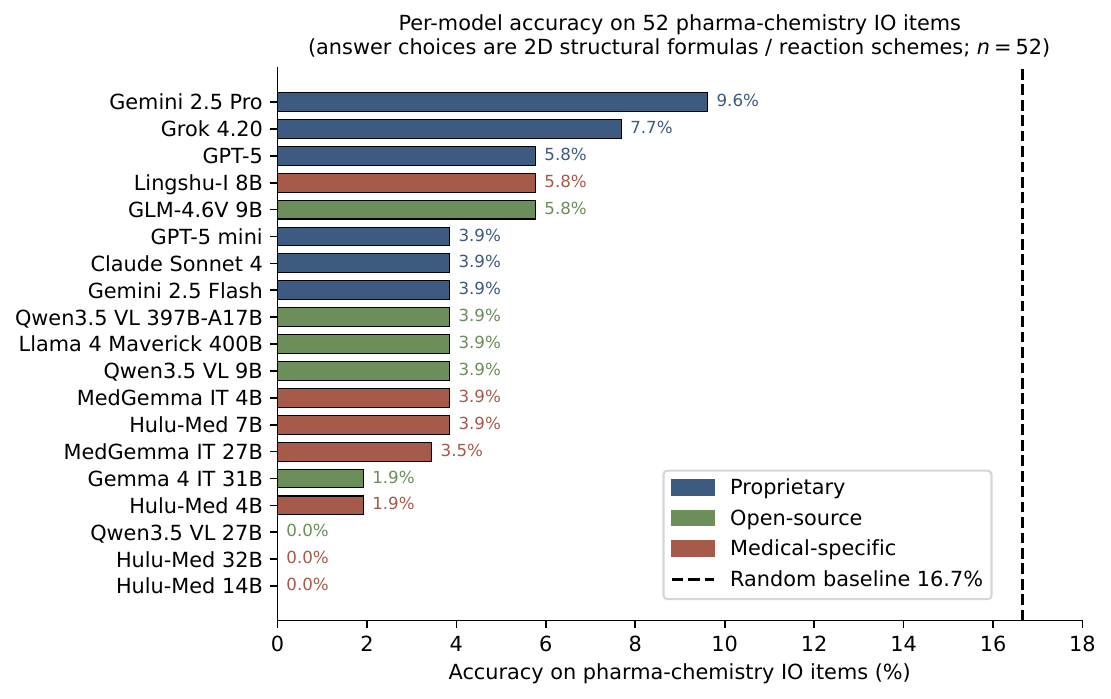}
  \caption{\textbf{Per-model accuracy on 52 pharma-chemistry IO items.}
  Dashed line marks the nominal random baseline $1/6=16.7\%$.  All 19
  models score below this baseline; the best score (Gemini 2.5 Pro, $9.6\%$)
  is still $7$\,pp below random.  Bar colours: blue = frontier, grey =
  mid open-source, red = medical-specific.}
  \label{fig:io-pharma}
\end{figure}

% app:io-pharma
Figure~\ref{fig:io-pharma} reports per-model accuracy on the 52
pharma-chemistry IO items introduced in \S\ref{app:image_options}.
Each item presents 2D structural formulas or reaction schemes as
the image options; no language prior can map the question text to
the correct chemical structure without reading the image.

All 19 models score below the $1/6 = 16.7\%$ random baseline, ranging
from $0.0\%$ to $9.6\%$.  The below-random pattern indicates that
strong language priors actively mislead: models confidently select the
most linguistically plausible option, which is systematically wrong when
the correct answer requires reading the chemical structure.  No
frontier-vs-medical separation appears on this subset: Lingshu-I 8B and
GLM-4.6V 9B tie with GPT-5 at $5.8\%$, above several frontier models.
The failure is universal---no model in our suite has reliable 2D
structural-formula reading capability.

%% file: sections/app_cases.tex
% Answer-transition case study transcripts
% 4 cases: Gemini 2.5 Pro, Physician NLE 2021
% Order: A, B, C, D (alphabetical)

\setlength{\fboxrule}{0.6pt}\setlength{\fboxsep}{4pt}

Each case presents the relevant image above a bilingual question stem,
the gold answer with both Japanese and English options, and side-by-side
Gemini 2.5 Pro predictions (baseline with image vs.\ vision ablation
without image).  Images are reproduced from the 2021 Japanese Physician
National Licensing Examination and are the property of the Ministry of
Health, Labour and Welfare of Japan.

\bigskip

%---A----------------------------------------------------------------
\paragraph{Case A\,: vision-induced error (Sect.\,F, Q\,52).}~\\[4pt]
\begin{center}
  \includegraphics[width=0.45\linewidth]{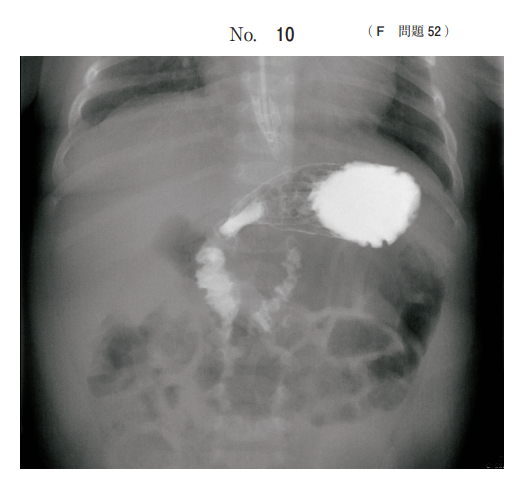}\\[2pt]
  \footnotesize\textit{Pediatric upper-GI contrast study (25-day-old female, vomiting workup).}
\end{center}

\noindent\small
\textbf{JP:} 日齢25の女児。嘔吐を主訴に来院。完全母乳栄養で生後11日頃から哺乳後に
1日2--3回の嘔吐を認める。吐物は母乳様、排便は毎日あり。身長50\,cm、体重3{,}520\,g
（14日前3{,}100\,g）。バイタル正常、軽度腹部膨満、肝・脾触知せず、
血液・画像検査に特記なし。上部消化管造影像（別冊No.\,10）を別に示す。
治療として適切なのはどれか。\\[3pt]
\textbf{EN:} A 25-day-old breastfed female with milk-coloured post-feed
vomiting (2--3 episodes/day) since day 11 and daily stools. Weight
3{,}520~g (was 3{,}100~g 14 days earlier); vital signs normal, mild
abdominal distension, otherwise unremarkable.  Upper-GI contrast study
shown. Which treatment is appropriate?\\[4pt]
\begin{tabular}{@{}l p{0.40\linewidth} p{0.42\linewidth}@{}}
   & \textit{JP} & \textit{EN} \\
  a & 緊急手術 & Emergency surgery\\
  b & 経管栄養 & Tube feeding\\
  c & 静脈栄養 & Parenteral nutrition\\
  d & 制吐薬投与 & Antiemetic\\
  \textbf{e} & \textbf{右側臥位保持}~~$\checkmark$ & \textbf{Right lateral positioning}~~$\checkmark$\\
\end{tabular}

\vspace{5pt}\noindent
\begin{tabular}{@{}p{0.47\linewidth}@{\hspace{5pt}}p{0.47\linewidth}@{}}
  \small\textbf{Baseline (image visible) --- a~$\times$} &
  \small\textbf{Vision ablation (no image) --- e~$\checkmark$}\\[2pt]
  \footnotesize ``上部消化管造影で十二指腸から空腸にかけて「コークスクリューサイン」を認める。腸回転異常症に伴う中腸軸捻転の典型所見であり、緊急手術（Ladd手術）の適応となる。'' &
  \footnotesize ``体重増加は良好（約30\,g/日）、脱水や電解質異常もなく、生理的な胃食道逆流（GER）が考えられる。第一選択は体位の工夫など保存的治療で、右側臥位保持が最も適切。'' \\[2pt]
  \multicolumn{2}{@{}p{\linewidth}@{}}{\footnotesize\textit{EN: Without the image the model uses the
  quantitative stem facts (weight gain $\approx$30\,g/day, daily stools, no dehydration) to
  recognise stable physiologic reflux and select positioning therapy.  With the image it anchors on
  a ``corkscrew sign'' reading from duodenum to jejunum, diagnoses midgut volvulus, and escalates
  to emergency Ladd surgery.}}\\
\end{tabular}

\noindent\fbox{~\textbf{A: vision-induced error}~}~\small
The image shifts the model away from a correct text-only answer.  The without-image reasoning leans
on quantitative stem facts (weight gain, daily stooling, normal labs) to recognise physiologic
reflux; with the image, anchoring on a corkscrew-sign reading triggers an emergency-surgery
recommendation, a clinically dramatic reversal of management.  Visual input actively misleads.

%---B----------------------------------------------------------------
\bigskip
\paragraph{Case B\,: text-solvable success (Sect.\,E, Q\,2).}~\\[4pt]
\begin{center}
  \includegraphics[width=0.55\linewidth]{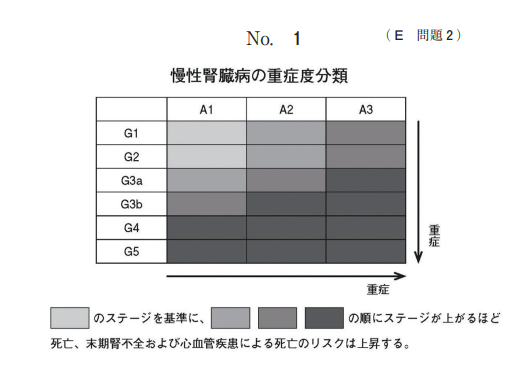}\\[2pt]
  \footnotesize\textit{CKD severity chart.}
\end{center}

\noindent\small
\textbf{JP:} 慢性腎臓病の重症度分類（別冊No.\,1）を別に示す。
A1からA3の方向（横軸方向）の区分を規定する指標はどれか。\\[3pt]
\textbf{EN:} The CKD severity classification is shown (figure).
Which indicator defines the A1--A3 horizontal-axis categories?\\[4pt]
\begin{tabular}{@{}l p{0.40\linewidth} p{0.42\linewidth}@{}}
   & \textit{JP} & \textit{EN} \\
  a & eGFR & eGFR\\
  b & 年齢 & Age\\
  \textbf{c} & \textbf{蛋白尿}~~$\checkmark$ & \textbf{Proteinuria/albuminuria}~~$\checkmark$\\
  d & 体格指数 & Body mass index\\
  e & 平均血圧 & Mean blood pressure\\
\end{tabular}

\vspace{5pt}\noindent
\begin{tabular}{@{}p{0.47\linewidth}@{\hspace{5pt}}p{0.47\linewidth}@{}}
  \small\textbf{Baseline (image visible) --- c~$\checkmark$} &
  \small\textbf{Vision ablation (no image) --- c~$\checkmark$}\\[2pt]
  \footnotesize ``GFR区分（縦軸）と蛋白尿A分類（横軸）を組み合わせて評価。A1からA3の区分は蛋白尿。'' &
  \footnotesize ``横軸は蛋白尿・アルブミン尿の程度を示し、A1（正常），A2（中等度），A3（高度）。'' \\[2pt]
  \multicolumn{2}{@{}p{\linewidth}@{}}{\footnotesize\textit{EN: Both runs invoke the standard textbook rule
  (CKD A-classification = proteinuria axis). The image is present but adds no new information.}}\\
\end{tabular}

\noindent\fbox{~\textbf{B: text-solvable success}~}~\small
The gold answer is recoverable from pre-training. Both conditions return the same correct
prediction with overlapping reasoning; image access is irrelevant.

%---C----------------------------------------------------------------
\bigskip
\paragraph{Case C\,: vision-enabled correction (Sect.\,B, Q\,8).}~\\[4pt]
\begin{center}
  \includegraphics[width=0.65\linewidth]{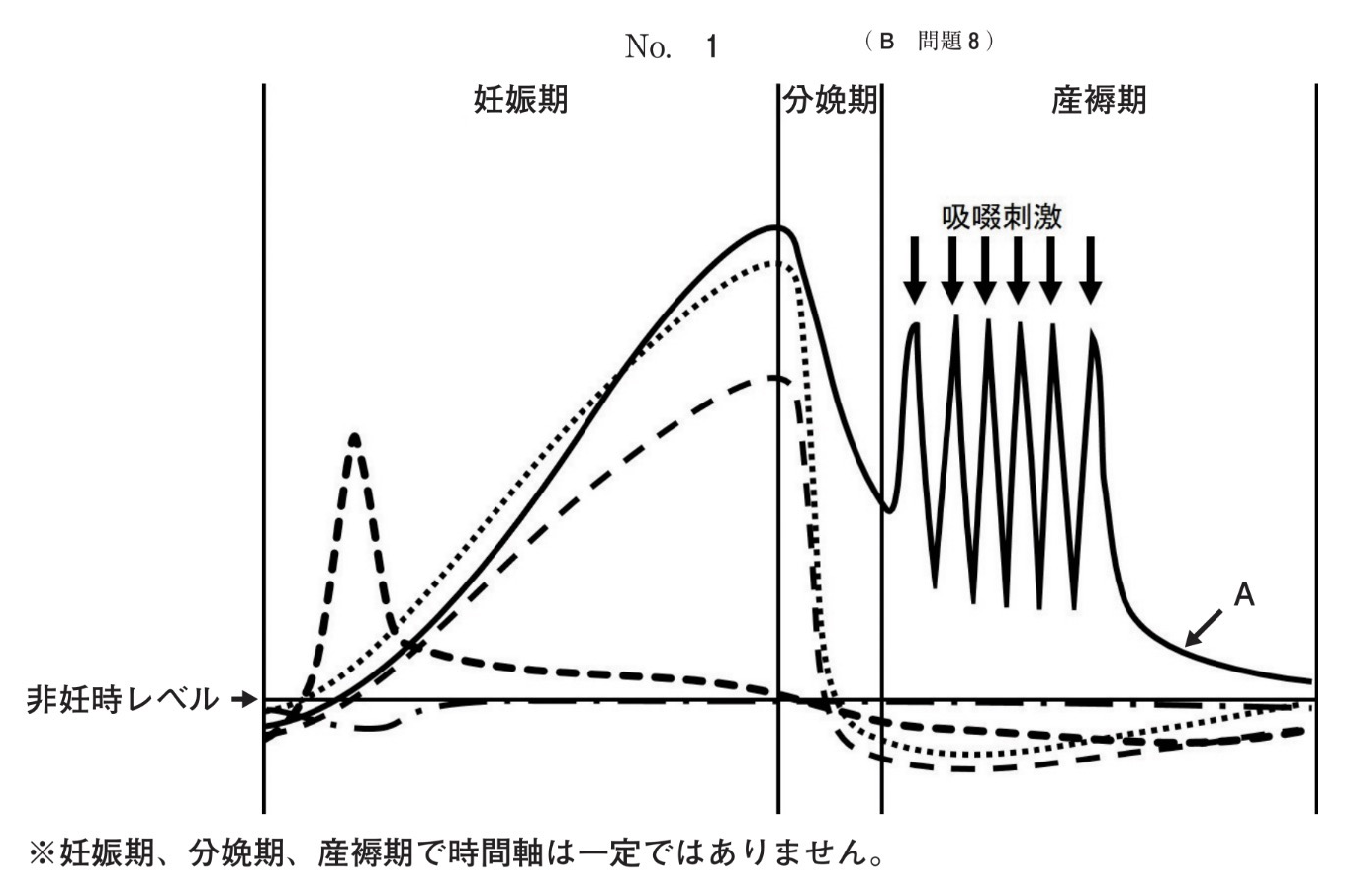}\\[2pt]
  \footnotesize\textit{Hormone curves; pulsatile post-partum peaks identify line~A as prolactin.}
\end{center}

\noindent\small
\textbf{JP:} 妊娠、分娩、産褥期における母体血中ホルモン値の変化（別冊No.\,1）を別に示す。
実線Aが表しているのはどれか。\\[3pt]
\textbf{EN:} Maternal blood hormone levels during pregnancy,
parturition, and the puerperium are shown.  Which hormone does
solid line~A represent?\\[4pt]
\begin{tabular}{@{}l p{0.40\linewidth} p{0.42\linewidth}@{}}
   & \textit{JP} & \textit{EN} \\
  a & エストロゲン & Estrogen\\
  \textbf{b} & \textbf{プロラクチン}~~$\checkmark$ & \textbf{Prolactin}~~$\checkmark$\\
  c & プロゲステロン & Progesterone\\
  d & 甲状腺刺激ホルモン & TSH\\
  e & 絨毛性ゴナドトロピン & hCG\\
\end{tabular}

\vspace{5pt}\noindent
\begin{tabular}{@{}p{0.47\linewidth}@{\hspace{5pt}}p{0.47\linewidth}@{}}
  \small\textbf{Baseline (image visible) --- b~$\checkmark$} &
  \small\textbf{Vision ablation (no image) --- e~$\times$}\\[2pt]
  \footnotesize ``産褥期の吸啜刺激によって急激な分泌が繰り返される実線Aのパターンはプロラクチンと一致する。'' &
  \footnotesize ``hCGは妊娠8〜10週頃にピークを迎え、その後減少する。実線Aはこの特徴的パターンと一致するため。'' \\[2pt]
  \multicolumn{2}{@{}p{\linewidth}@{}}{\footnotesize\textit{EN: With the image the model reads the
  pulsatile post-partum surges (unique to prolactin) and answers correctly.  Without the image it
  defaults to the hCG bell-curve---the most statistically prominent hormone arc in its training
  distribution---which is incorrect.}}\\
\end{tabular}

\noindent\fbox{~\textbf{C: vision-enabled correction}~}~\small
The image provides the decisive evidence.  Post-partum pulsatile surges are unique to prolactin
and cannot be inferred from the text stem; removing the image activates an incorrect language
prior (hCG bell-curve).

%---D----------------------------------------------------------------
\bigskip
\paragraph{Case D\,: model-limit failure (Sect.\,F, Q\,17).}~\\[4pt]
\begin{center}
  \includegraphics[width=0.55\linewidth]{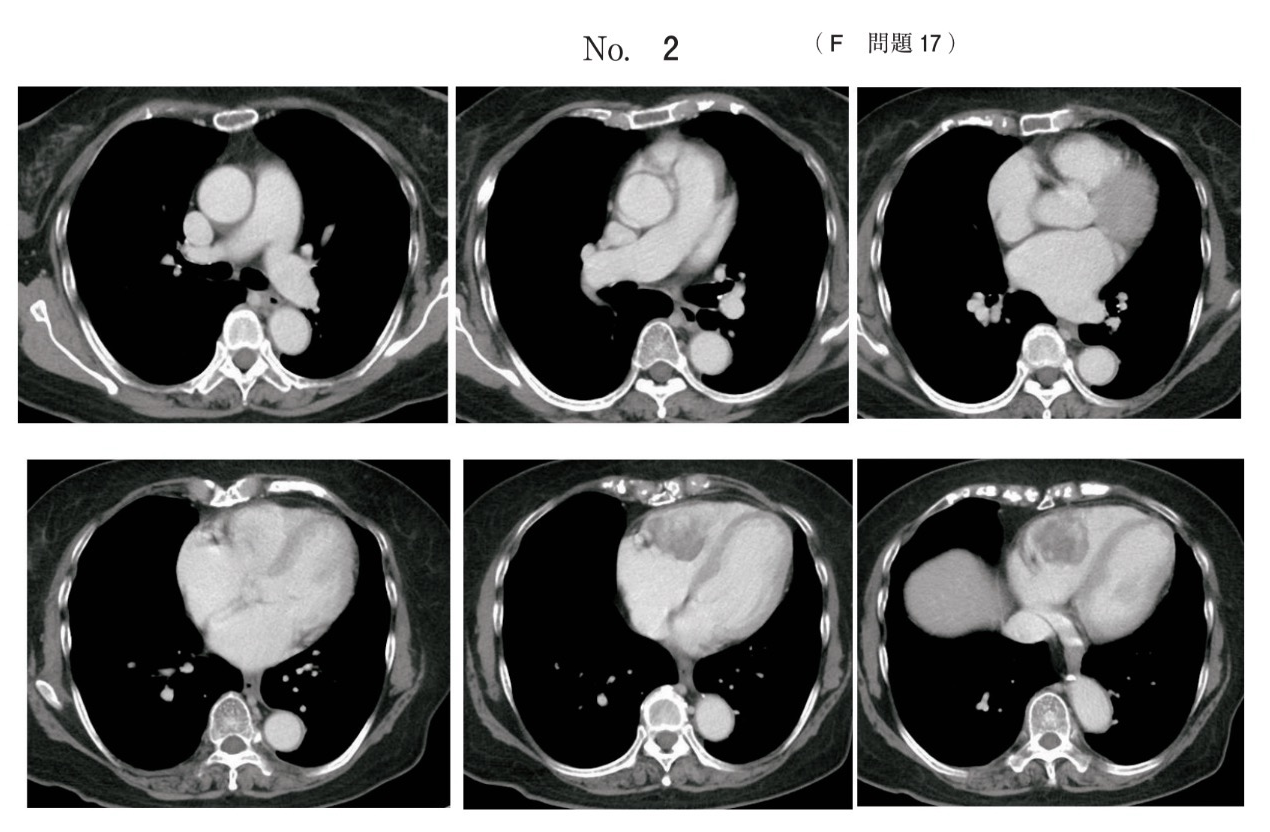}\\[2pt]
  \footnotesize\textit{Mediastinal CT; right ventricle (anterior) is enlarged.}
\end{center}

\noindent\small
\textbf{JP:} 縦隔条件の胸部造影CT（別冊No.\,2）を別に示す。
異常所見を示す心臓の部位はどこか。\\[3pt]
\textbf{EN:} A mediastinal-window contrast CT of the chest is shown.
Which cardiac structure shows an abnormality?\\[4pt]
\begin{tabular}{@{}l p{0.40\linewidth} p{0.42\linewidth}@{}}
   & \textit{JP} & \textit{EN} \\
  a & 右房 & Right atrium\\
  b & 左房 & Left atrium\\
  c & 肺動脈 & Pulmonary artery\\
  \textbf{d} & \textbf{右心室}~~$\checkmark$ & \textbf{Right ventricle}~~$\checkmark$\\
  e & 左心室 & Left ventricle\\
\end{tabular}

\vspace{5pt}\noindent
\begin{tabular}{@{}p{0.47\linewidth}@{\hspace{5pt}}p{0.47\linewidth}@{}}
  \small\textbf{Baseline (image visible) --- a~$\times$} &
  \small\textbf{Vision ablation (no image) --- b~$\times$}\\[2pt]
  \footnotesize ``右心房内に造影効果の乏しい充満欠損が認められる。'' &
  \footnotesize ``心臓の最も後方に位置する左房が著明に拡大している。'' \\[2pt]
  \multicolumn{2}{@{}p{\linewidth}@{}}{\footnotesize\textit{EN: The model produces a different wrong
  answer in each condition---right atrium with the image, left atrium without.  Neither matches the
  correct answer (right ventricle).  The image perturbs output direction without correcting it.}}\\
\end{tabular}

\noindent\fbox{~\textbf{D: model-limit failure}~}~\small
The model fails regardless of image availability and assigns a different structure in each
condition.  Neither visual input nor language prior provides sufficient signal to identify right
ventricular enlargement; image access shifts but does not resolve the model's uncertainty.

%% file: checklist.tex
\section*{NeurIPS Paper Checklist}

\begin{enumerate}

\item {\bf Claims}
    \item[] Question: Do the main claims made in the abstract and introduction accurately reflect the paper's contributions and scope?
    \item[] Answer: \answerYes{}
    \item[] Justification: The abstract and the contributions list in Section~\ref{sec:intro} enumerate the JMed48k corpus and JMed48k-Eval evaluation subset, the systematic 21-model evaluation across proprietary, open-source, and medical-specific cohorts, and the paired image-removal audit, all of which are realised in Sections~\ref{sec:dataset}--\ref{sec:results} and the supporting appendices.

\item {\bf Limitations}
    \item[] Question: Does the paper discuss the limitations of the work performed by the authors?
    \item[] Answer: \answerYes{}
    \item[] Justification: In Section~\ref{sec:conclusion}.

\item {\bf Theory assumptions and proofs}
    \item[] Question: For each theoretical result, does the paper provide the full set of assumptions and a complete (and correct) proof?
    \item[] Answer: \answerNA{}
    \item[] Justification: The paper makes no theoretical claims; its contribution is an empirical multi-profession licensing benchmark and a paired image-removal audit, so no theorems, formal assumptions, or proofs are required.

    \item {\bf Experimental result reproducibility}
    \item[] Question: Does the paper fully disclose all the information needed to reproduce the main experimental results of the paper to the extent that it affects the main claims and/or conclusions of the paper (regardless of whether the code and data are provided or not)?
    \item[] Answer: \answerYes{}
    \item[] Justification: Appendix~\ref{app:setup} documents the 21 model snapshots, cohort partition, and inference protocol (provider APIs for closed-source; vLLM with fixed seeds, temperature~0, single-shot Japanese prompts, and per-model retry policies for open-source); Appendix~\ref{app:setup_prompts} gives both prompt templates verbatim; and Appendix~\ref{app:scoring} specifies the deterministic answer-normalisation and refusal/parse-failure rules used to compute every reported number on the fixed $12{,}484$-item scored subset.

\item {\bf Open access to data and code}
    \item[] Question: Does the paper provide open access to the data and code, with sufficient instructions to faithfully reproduce the main experimental results, as described in supplemental material?
    \item[] Answer: \answerYes{}
    \item[] Justification: Link is in Abstract.

\item {\bf Experimental setting/details}
    \item[] Question: Does the paper specify all the training and test details (e.g., data splits, hyperparameters, how they were chosen, type of optimizer) necessary to understand the results?
    \item[] Answer: \answerYes{}
    \item[] Justification: Appendix~\ref{app:setup} reports all training-free inference details (model snapshots, vLLM configuration, decoding parameters, retry policy, image-removed prompt construction), and Appendix~\ref{app:scoring} specifies the scoring contract that defines what counts as a correct prediction across single-key, multi-key, alternative-accepting, and numeric-slot answer formats.

\item {\bf Experiment statistical significance}
    \item[] Question: Does the paper report error bars suitably and correctly defined or other appropriate information about the statistical significance of the experiments?
    \item[] Answer: \answerNo{}
    \item[] Justification: All reported numbers are exact item-level accuracies on a fixed shared scored denominator of $12{,}484$ items; we deliberately use a single-pass deterministic protocol (temperature~0, fixed seeds) on a fixed question set, so each (model, profession) cell is evaluated once and no error bars, confidence intervals, or hypothesis tests are reported.

\item {\bf Experiments compute resources}
    \item[] Question: For each experiment, does the paper provide sufficient information on the computer resources (type of compute workers, memory, time of execution) needed to reproduce the experiments?
    \item[] Answer: \answerNo{}
    \item[] Justification: The paper does not characterise GPU type, memory footprint, wall-clock time, or provider API cost; closed-source models were queried through provider APIs at the snapshots listed in Appendix~\ref{app:setup}, and open-source models were run on local vLLM at temperature~0 with fixed seeds, but the underlying hardware is not reported in this version.

\item {\bf Code of ethics}
    \item[] Question: Does the research conducted in the paper conform, in every respect, with the NeurIPS Code of Ethics \url{https://neurips.cc/public/EthicsGuidelines}?
    \item[] Answer: \answerYes{}
    \item[] Justification: The work uses only public materials officially released by the Japanese Ministry of Health, Labour and Welfare, conducts no human-subjects research, and respects the NeurIPS Code of Ethics; the privacy-driven exclusion of image content that MHLW does not itself redistribute is documented in Section~\ref{sec:dataset-taxonomy}.

\item {\bf Broader impacts}
    \item[] Question: Does the paper discuss both potential positive societal impacts and negative societal impacts of the work performed?
    \item[] Answer: \answerYes{}
    \item[] Justification: The paper discusses construct validity (licensing-exam performance is not equivalent to clinical-deployment readiness; see Section~\ref{sec:conclusion}), data-contamination considerations on public exam material (Appendix~\ref{app:image_options}), and the behavioural-rather-than-mechanistic interpretation of the paired image-removal audit (Appendix~\ref{app:refusal}), all of which bear on responsible interpretation and downstream use of the benchmark.

\item {\bf Safeguards}
    \item[] Question: Does the paper describe safeguards that have been put in place for responsible release of data or models that have a high risk for misuse (e.g., pre-trained language models, image generators, or scraped datasets)?
    \item[] Answer: \answerYes{}
    \item[] Justification: The corpus is built from official MHLW exam releases and never redistributes image content that MHLW does not itself publish (typically patient portraits or similar privacy-sensitive material); such references are isolated in an \emph{Other/Unclear} bucket and excluded from type-stratified analyses, as documented in Section~\ref{sec:dataset-taxonomy}.

\item {\bf Licenses for existing assets}
    \item[] Question: Are the creators or original owners of assets (e.g., code, data, models), used in the paper, properly credited and are the license and terms of use explicitly mentioned and properly respected?
    \item[] Answer: \answerYes{}
    \item[] Justification: All third-party assets used in the paper---the MinerU layout-extraction system (Section~\ref{sec:dataset-construction}), the 21 evaluated proprietary and open-source vision-language models (Appendix~\ref{app:setup}), and prior licensing benchmarks cited in Section~\ref{sec:related}---are credited via citations and used under their original public terms; the source examination materials are official MHLW releases used under Japanese government open-data terms, and JMed48k is planned for release as a derivative dataset under a permissive open license (e.g., CC~BY~4.0) compatible with those terms.

\item {\bf New assets}
    \item[] Question: Are new assets introduced in the paper well documented and is the documentation provided alongside the assets?
    \item[] Answer: \answerYes{}
    \item[] Justification: Documentation for the JMed48k release ships with the paper: the released JSON schema with field-by-field descriptions (Appendix~\ref{app:curation_schema}, Table~\ref{tab:schema_fields}), the PDF-to-JSON construction and quality-control pipeline (Appendix~\ref{app:curation}), the 8-type image taxonomy and its annotation procedure (Appendix~\ref{app:image_taxonomy}), and the deterministic scoring contract (Appendix~\ref{app:scoring}).

\item {\bf Crowdsourcing and research with human subjects}
    \item[] Question: For crowdsourcing experiments and research with human subjects, does the paper include the full text of instructions given to participants and screenshots, if applicable, as well as details about compensation (if any)?
    \item[] Answer: \answerNA{}
    \item[] Justification: The paper involves no crowdsourcing or human-subjects research; the only human involvement is a professional Japanese-language data-correction vendor performing quality-control review of the PDF-to-JSON extraction (Appendix~\ref{app:curation_qc}), not data generation, annotation crowdsourcing, or participant elicitation.

\item {\bf Institutional review board (IRB) approvals or equivalent for research with human subjects}
    \item[] Question: Does the paper describe potential risks incurred by study participants, whether such risks were disclosed to the subjects, and whether Institutional Review Board (IRB) approvals (or an equivalent approval/review based on the requirements of your country or institution) were obtained?
    \item[] Answer: \answerNA{}
    \item[] Justification: No human-subjects research is conducted; the data are derived from publicly released MHLW examination materials, so no IRB or equivalent ethics-board review applies.

\item {\bf Declaration of LLM usage}
    \item[] Question: Does the paper describe the usage of LLMs if it is an important, original, or non-standard component of the core methods in this research? Note that if the LLM is used only for writing, editing, or formatting purposes and does \emph{not} impact the core methodology, scientific rigor, or originality of the research, declaration is not required.
    %this research?
    \item[] Answer: \answerYes{}
    \item[] Justification: A frontier vision--language model is used as an intermediate structuring tool, in conjunction with MinerU layout extraction, to convert MHLW question, figure, and answer PDFs into the released JSON records (Appendix~\ref{app:curation}); gold answers are aligned separately and verbatim from the official MHLW answer keys and are never produced by any model, and the structured outputs pass automated consistency checks, human annotator review, and a professional Japanese-language data-correction vendor pass before release.

\end{enumerate}